\documentclass[11pt]{article}

\usepackage[final]{acl}

\usepackage{times}
\usepackage{latexsym}
\usepackage{stfloats}
\usepackage[T1]{fontenc}

\usepackage[utf8]{inputenc}

\usepackage{microtype}

\usepackage{inconsolata}

\usepackage{graphicx}

\usepackage{graphicx}
\usepackage{booktabs} 
\usepackage{multirow}
\usepackage{adjustbox}
\usepackage{float} 
\usepackage{colortbl}
\usepackage{xcolor}
\usepackage{longtable}
\usepackage{arydshln}
\usepackage{pifont}
\usepackage{array}
\usepackage[utf8]{inputenc}
\usepackage{subcaption}
\usepackage{algorithm}
\usepackage{algorithmic}
\usepackage{listings}
\usepackage{wrapfig}  
\usepackage{enumitem}
\usepackage{arydshln}
\usepackage{hyperref} 

\usepackage{multicol}
\usepackage[skins,breakable]{tcolorbox}
\usepackage[dvipsnames,svgnames,x11names]{xcolor}
\usepackage{varwidth}
\usepackage{makecell}

\usepackage{tikz}
\usepackage[skins]{tcolorbox}
\usepackage{listings}

\usepackage{amsmath}
\usepackage{amssymb}
\tcbuselibrary{breakable, skins}

\title{TRUST-SQL: Tool-Integrated Multi-Turn Reinforcement Learning for Text-to-SQL over Unknown Schemas}

\author{
	Ai Jian$^{1}$\thanks{\ \ Equal contribution; \textsuperscript{$\dagger$} Corresponding author; this work was conducted during an internship at Meituan.}, Xiaoyun Zhang$^{2,3\ast}$, Eryu guo$^{1}$, Wanrou Du$^{1}$, Jingqing Ruan$^{4\dagger}$\\
	\textbf{Jiangbo Pei$^{1}$, Weipeng Zhang}$^{4}$, \textbf{Ke Zeng}$^{4}$, \textbf{Xunliang Cai}$^{4}$ \\
	$^{1}$Beijing University of Posts and Telecommunications, Beijing, China \\
	$^{2}$State Key Lab of Processors, Institute of Computing Technology, CAS \\
	$^{3}$University of Chinese Academy of Sciences, $^{4}$Meituan, Beijing, China \\
	\texttt{jianai@bupt.edu.cn}, \texttt{ruanjingqing@meituan.com}
}

\begin{document}
\maketitle
\begin{abstract}

Text-to-SQL parsing has achieved remarkable progress under the \textbf{Full Schema Assumption}, where the complete database structure is pre-loaded into the model's context. Yet this assumption overlooks a more fundamental capability question of whether an LLM agent can resolve a query without explicit access to the underlying tables and relations. Moreover, full-schema injection may exceed the effective context window, introduce irrelevant metadata into the reasoning process, and become less reliable as databases evolve.
We therefore study the \textbf{Unknown Schema}
setting, in which an agent must actively explore the
database and commit to only the relevant subset of metadata.
To address this, we propose \textbf{TRUST-SQL}
(\textbf{T}ruthful \textbf{R}easoning with \textbf{U}nknown
\textbf{S}chema via \textbf{T}ools). We formulate the task
as a \textbf{Partially Observable Markov Decision Process}
in which our autonomous agent follows a structured
\textbf{four-phase protocol} to ground reasoning in
verified metadata. This protocol provides a
structural boundary for our novel \textbf{Phase-Aware GRPO}
strategy, which applies token-level masked advantages to
isolate exploration rewards from execution outcomes and
resolve credit assignment, yielding a \textbf{9.9\%
relative improvement} over standard GRPO.
Extensive
experiments across five benchmarks demonstrate that
TRUST-SQL is effective at \textbf{multiple scales (4B/ 8B/14B)}.
Despite
operating entirely without pre-loaded metadata, our agent
\textbf{matches or surpasses prior baselines on most benchmarks}
that rely on schema prefilling.\footnote{Data, code, and checkpoints are available at \url{https://huggingface.co/collections/AIJian/trustsql}.}
\end{abstract}

\section{Introduction}

Text-to-SQL parsing, which translates natural language questions into executable SQL queries, has seen remarkable progress driven by Large Language Models (LLMs)~\citep{metadata, scale}. However, much of this progress relies on a critical yet underexplored premise known as the \textbf{Full Schema Assumption}, where the complete database structure is pre-loaded into the model's context. Under this setting, the task is largely framed as a \textit{static translation} problem, and existing methods report strong performance on benchmarks with pre-injected database structures~\citep{bird, spider}.

\begin{figure}[t]
	\centering
	\includegraphics[width=0.5\textwidth]{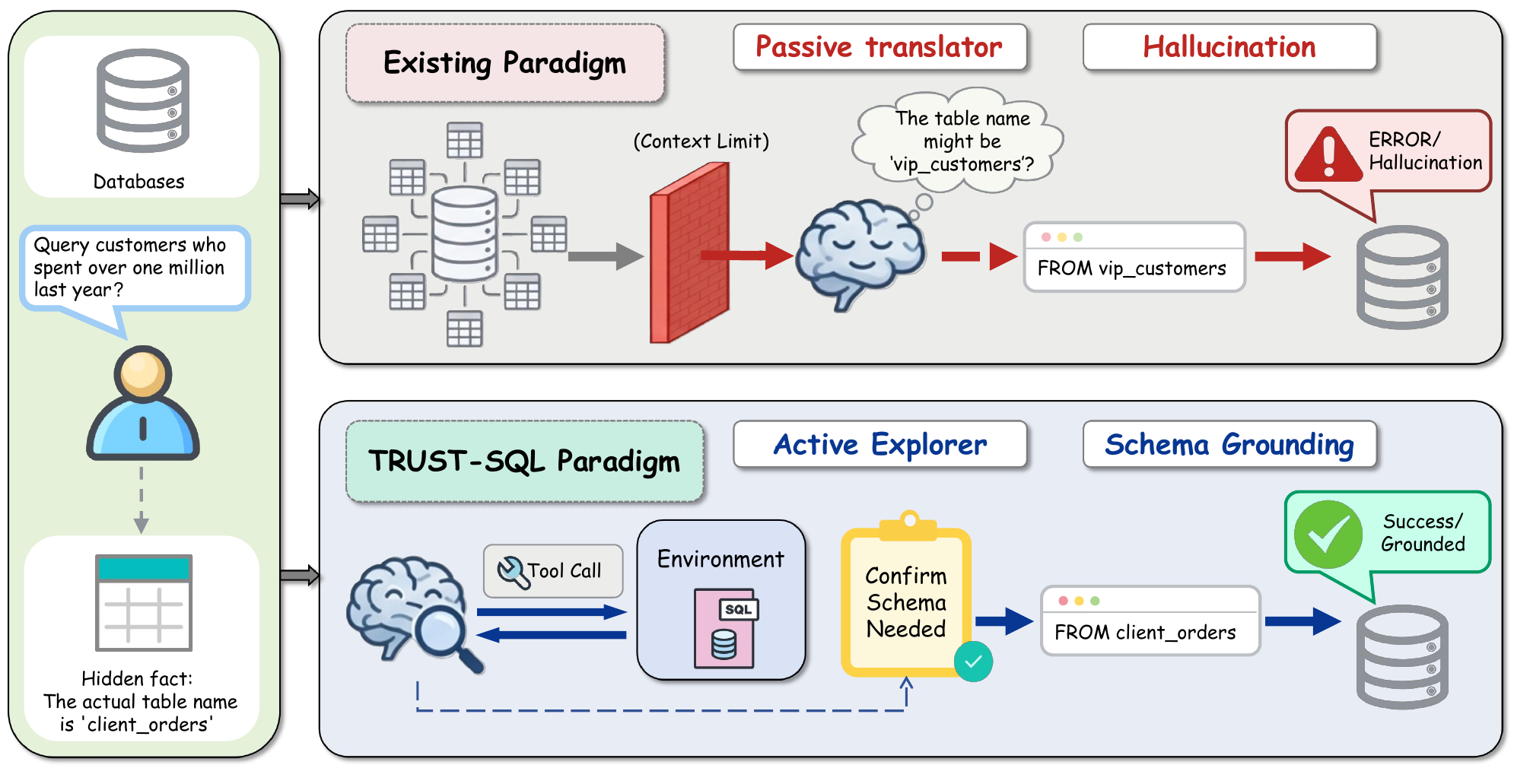}
	\caption{Existing methods rely on pre-loaded schemas, while the Unknown Schema setting requires exploration.}
	\label{fig:paradigm_comparison}
	\vspace{-4mm}
\end{figure}

This assumption is both empirically fragile and conceptually restrictive, as full-schema injection introduces substantial irrelevant metadata into the prompt as database structures grow, increasing contextual burden and potentially degrading reasoning quality, with important information often buried in the middle of the prompt~\citep{lost}.
 Meanwhile, real-world databases continuously evolve through additions, deletions, and structural modifications~\citep{evoschema}, making static representations increasingly incomplete over time. More importantly, this paradigm avoids a deeper capability question of whether an LLM agent can resolve a database query without direct access to the full database structure. As illustrated in Figure~\ref{fig:paradigm_comparison}, we define this missing setting as the \textbf{Unknown Schema} scenario, where an agent must move beyond passive consumption and actively explore the database to retrieve only the metadata necessary for the query.

Once the schema is no longer pre-loaded, standard single-turn methods break down, as they lack any mechanism for active retrieval under partial observability. The task must instead be formulated as a \textbf{multi-turn tool-integrated decision-making process}. Recent agentic frameworks have explored this iterative setting, yet two key bottlenecks remain unresolved.
\textbf{\textit{Architecturally}}, LLMs struggle to maintain coherent reasoning over long interaction horizons. Without explicit grounding mechanisms, they often lose track of intermediate observations~\citep{lost} and fall back on hallucinating schema elements from parametric priors.
\textbf{\textit{Algorithmically}}, credit assignment across long interaction trajectories remains a fundamental challenge for large language models~\citep{sweet_rl, duca}. Existing approaches typically rely on a single terminal reward~\citep{agro,mtir} or naive aggregation of intermediate signals~\citep{sqltrail}, which conflates schema exploration quality with SQL generation quality and makes it difficult to attribute final execution outcomes to specific actions.

In this paper, we propose \textbf{TRUST-SQL} (\textbf{T}ruthful \textbf{R}easoning with \textbf{U}nknown \textbf{S}chema via \textbf{T}ools) to systematically address these challenges.
We formulate the task as a Partially Observable Markov Decision Process and introduce a \textbf{four-phase interaction protocol} comprising \textit{Explore}, \textit{Propose}, \textit{Generate}, and \textit{Confirm}.
The \textit{Propose} phase acts as a mandatory cognitive
checkpoint that forces the agent to commit to verified
metadata before generation, suppressing downstream
hallucination. This checkpoint also provides a
structural boundary for \textbf{Phase-Aware GRPO}, a
training strategy built upon Group Relative Policy
Optimization (GRPO)~\citep{deepseek_r1} that applies
token-level masked advantages to separately credit schema
exploration and SQL execution, co-optimizing schema
grounding and SQL generation.

Our contributions are summarized as follows:

\begin{itemize}[leftmargin=*, topsep=3pt, itemsep=0pt]
    \item We formalize the \textbf{Unknown Schema} setting as a partial-observability testbed for agentic Text-to-SQL, and propose \textbf{TRUST-SQL}, an autonomous framework that interacts directly with unobservable databases to retrieve and verify metadata, supporting grounded SQL generation via iterative exploration instead of static context.
    
    \item We propose \textbf{Phase-Aware GRPO}, a novel training
    strategy utilizing token-level masked advantages and
    execution-coupled schema rewards. This granular optimization
    yields a \textbf{9.9\% relative improvement} in execution accuracy
    over standard GRPO on BIRD-Dev.

    \item Extensive experiments across five diverse
    benchmarks demonstrate that TRUST-SQL is \textbf{consistently
    effective across the 4B, 8B, and 14B scales}. Despite
    operating without pre-loaded metadata, our models match
    or surpass baselines that rely on schema injection on
    most benchmarks.
\end{itemize}

\section{Related Work}
\label{sec:related_work}

\noindent\textbf{Text-to-SQL under Full Schema Assumption.}
Most existing methods operate under the premise of full
schema observability. Supervised fine-tuning approaches
such as OmniSQL~\citep{omnisql}, STAR~\citep{star}, and
ROUTE~\citep{route} internalize generation capabilities
but rely on static context. Similarly, single-turn
reinforcement learning~(RL) methods~\citep{sql_r1, arctic, reward_sql, reasoning_sql}
optimize execution accuracy using terminal rewards while
assuming access to the complete database structure.
Constrained to a single-turn interaction paradigm, these models function as passive translators and struggle under Unknown Schema setting requiring active database exploration.

\noindent\textbf{Tool-Augmented Approaches for Text-to-SQL.}
To handle complex or hidden databases, recent works introduce tool-augmented methods. Training-free frameworks~\citep{toolsql,macsql,autolink} leverage frozen language models to query metadata, but remain susceptible to parametric hallucinations and lack reliable verification mechanisms. More recently, multi-turn RL approaches~\citep{mtir, sqltrail,mtsql} incorporate SQL execution into the training loop to refine queries. 
While promising, these methods lack clear separation between metadata retrieval and SQL generation, relying on conflated terminal rewards that fail to distinguish errors from schema retrieval and SQL generation.

\noindent\textbf{Credit Assignment in Multi-Turn RL.}
A central challenge in multi-turn RL is attributing final outcomes to individual actions in long trajectories. Existing solutions explore trajectory-level optimization~\citep{ragen, simpletir}, reward decomposition~\citep{epo}, tree-structured search~\citep{tree}, and intrinsic motivation~\citep{score, curio}. These techniques are primarily designed for homogeneous action spaces where each step contributes similarly to the final objective. However, Text-to-SQL introduces a heterogeneous decision process where schema retrieval and SQL generation play distinct roles, making standard credit assignment formulations insufficient. TRUST-SQL addresses this by introducing Phase-Aware GRPO to disentangle optimization across phases.

\begin{figure*}[t]
	\centering
	\includegraphics[width=\textwidth]{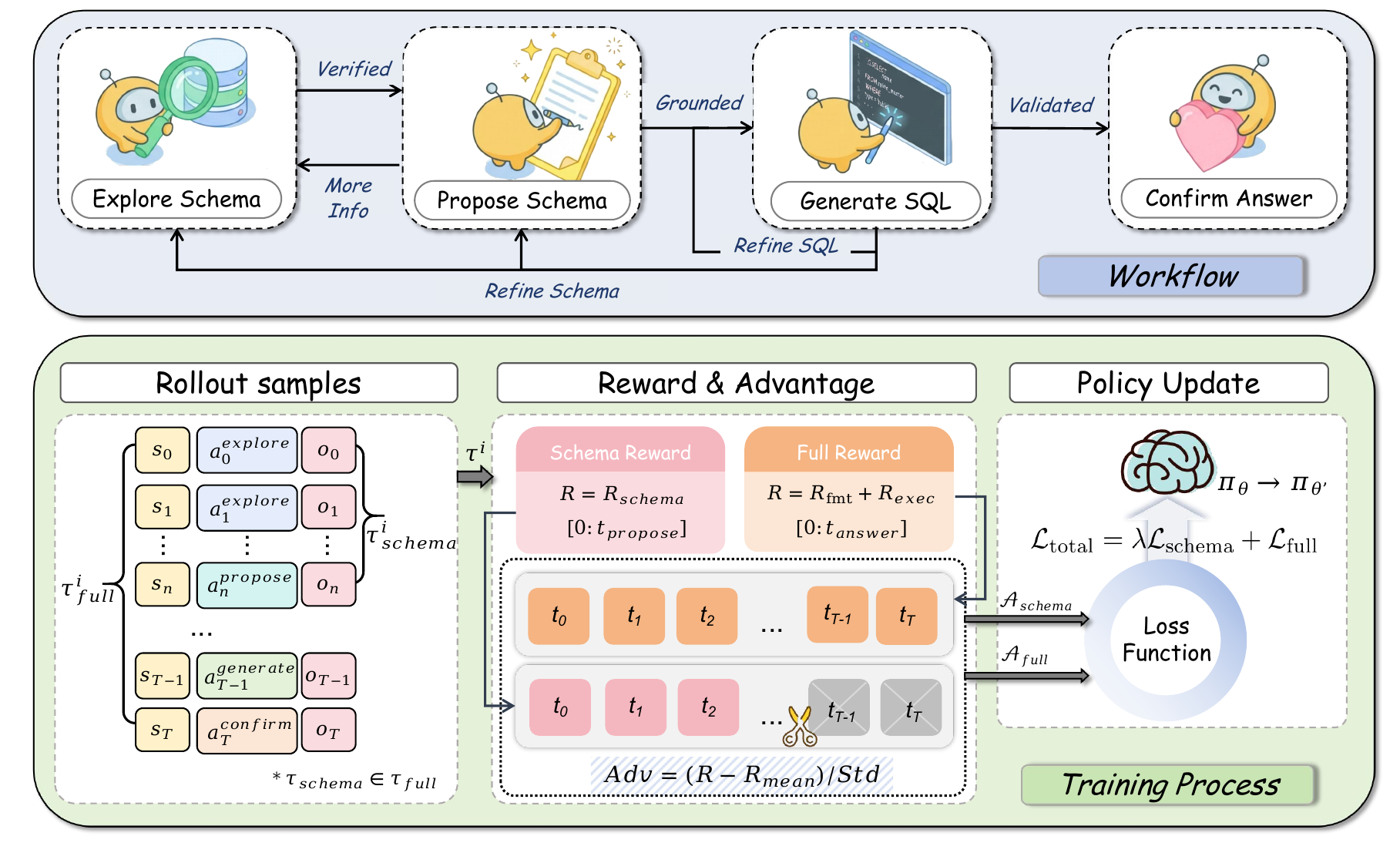}
	\caption{Overview of the \textbf{TRUST-SQL} framework. 
		\textbf{(Top)} The four-phase workflow comprising 
		\textit{Explore}, \textit{Propose}, \textit{Generate}, and 
		\textit{Confirm}, with non-linear transitions enabling 
		iterative schema refinement. \textbf{(Bottom)} The Phase-Aware
		GRPO training pipeline, where trajectories are decomposed into
		a Schema Track $\tau_{\text{schema}}$ and a Full Track
		$\tau_{\text{full}}$, each optimized with independent rewards 
		and masked advantages.}
	\label{fig:method}
\end{figure*}

\section{Methodology}
\label{sec:methodology}

We present \textbf{TRUST-SQL} for Text-to-SQL over unknown schemas. As illustrated in Figure~\ref{fig:method}, it consists of a four-phase interaction protocol and Phase-Aware GRPO training strategy. We formulate the task as a sequential decision-making process, followed by reward design and RL optimization.

\subsection{Motivation: Why a Four-Phase Protocol?}
\label{subsec:motivation}

To empirically justify our interaction protocol design and identify key bottlenecks of Text-to-SQL under the \textbf{Unknown Schema setting}, we conduct a pilot study on the \textbf{BIRD-Dev} dataset with \textbf{Qwen3-8B} as the base model. We construct \textbf{three agent variants} with incremental structural constraints on interaction behavior and classify failure cases to derive design principles for our framework.

\noindent\textbf{Protocol Variants.}
\textbf{EC} (Explore-Confirm) is a minimal baseline where 
the agent freely queries metadata and directly submits a SQL 
answer without intermediate verification. \textbf{EGC} 
(Explore-Generate-Confirm) introduces an \textbf{explicit 
Generate phase}, requiring the agent to execute a 
candidate SQL and observe its result before finalizing. 
\textbf{EPGC} (Explore-Propose-Generate-Confirm) further 
adds \textbf{the Propose phase} as a mandatory cognitive 
checkpoint, compelling the agent to commit to a verified 
schema before SQL generation.

\noindent\textbf{Error Taxonomy.}
We classify failures into five categories:
\textbf{(1) Hallucination}: the model fabricates 
non-existent tables or columns based on parametric priors;
\textbf{(2) Schema Linking}: the model selects wrong or 
missing tables and columns despite correct exploration;
\textbf{(3) Semantic}: the model correctly identifies the 
relevant schema but generates logically incorrect SQL;
\textbf{(4) Syntax}: the SQL contains malformed statements 
that fail to execute;
\textbf{(5) Generation}: the agent fails to produce a 
complete SQL, typically due to reaching the maximum turn 
limit.

\begin{figure}[t]
    \centering
    \begin{subfigure}[b]{0.65\columnwidth}
        \centering
        \includegraphics[width=\linewidth]{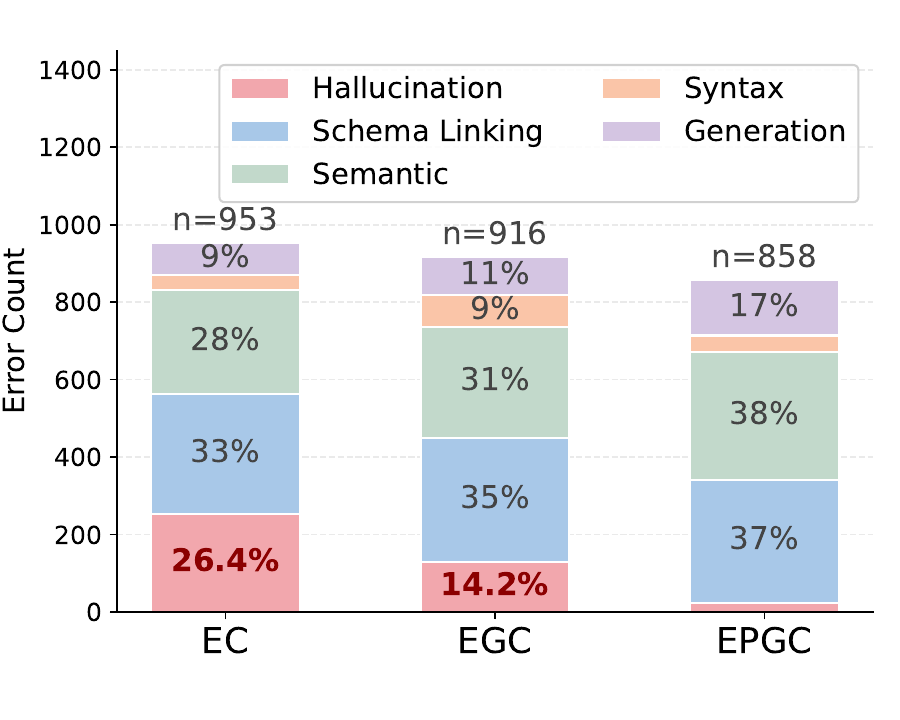}
        \caption{Stacked error distribution. }
        \label{fig:pilot_study_error}
    \end{subfigure}
    \hfill
    \begin{subfigure}[b]{0.33\columnwidth}
        \centering
        \includegraphics[width=\linewidth]{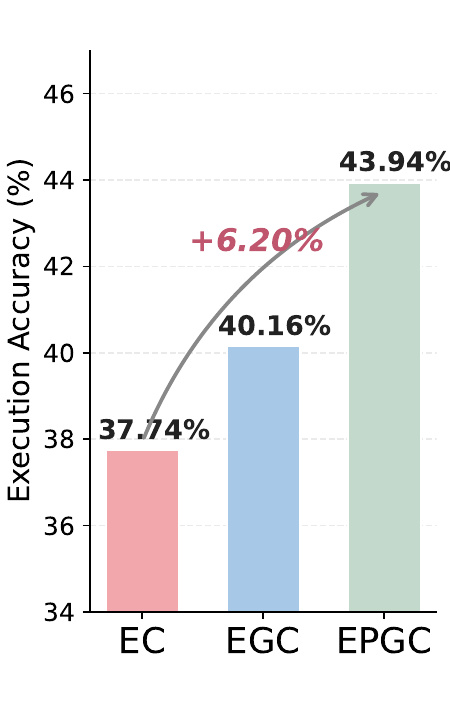}
        \caption{EX accuracy.}
        \label{fig:pilot_study_ex}
    \end{subfigure}
    \caption{Pilot study results on BIRD-Dev (Qwen3-8B).}
    \label{fig:pilot_study}
    \vspace{-3mm}
\end{figure}

As shown in Figure~\ref{fig:pilot_study}, three observations 
emerge from the results.

\noindent\textbf{Obs. 1:~Schema verification is critical to suppress hallucination.}
In EC, hallucination accounts for 26.4\% of all failures. 
The \textit{Generate} phase partially alleviates 
this via execution feedback (14.2\%), but the most 
significant reduction occurs with the \textit{Propose} in EPGC, driving hallucination to just 2.8\%, 
a \textbf{9.4$\times$ reduction} over EC.

\noindent\textbf{Obs. 2: Schema linking is the persistent 
	bottleneck.}
Schema linking errors remain consistently high across all 
variants, motivating our Phase-Aware
GRPO to provide an independent optimization signal for
schema exploration.

\noindent\textbf{Obs. 3: Suppressing hallucination reveals 
	semantic errors.}
As hallucination decreases, semantic errors increase from 
268 to 330, reflecting a distributional shift: once schema 
is correctly identified, complex query logic becomes the 
dominant challenge, motivating joint optimization of schema 
grounding and SQL generation.

These observations motivate the core designs of
our work: the \textit{Propose} checkpoint to suppress
hallucination, and Phase-Aware GRPO to co-optimize schema
exploration and SQL generation. The \textit{Propose}
checkpoint contributes at two levels. At the workflow
level, it grounds the agent's reasoning in verified
metadata. At the optimization level, it provides a
credit-assignment boundary for Phase-Aware GRPO. We
disentangle these two effects in
Appendix~\ref{app:propose_roles}.

\subsection{Problem Formulation}
\label{subsec:problem_formulation}

Based on the EPGC protocol validated in Section~\ref{subsec:motivation}, 
we formalize the Text-to-SQL task under the Unknown Schema 
setting as a Partially Observable Markov Decision Process 
\textbf{(POMDP)}, which is defined as $(\mathcal{S}, \mathcal{A}, \mathcal{T}, \mathcal{R}, \Omega, 
\mathcal{Z}, \gamma)$ over discrete steps $t = 0, 1, \dots, T$. 

\vspace{1mm}
\noindent\textbf{State and Observation Spaces.}
The true environment state $s_t \in \mathcal{S}$ represents 
the complete database schema and remains hidden from the 
agent. Consequently, the agent only receives partial 
observations $o_t \in \Omega$ dictated by the observation 
function $\mathcal{Z}$, which consist of tool execution 
feedback. To navigate this unobservable environment, the 
agent relies on an internal context state $c_t = (q, h_t, 
\mathcal{K}_t)$. This context integrates the user question 
$q$, the interaction history $h_t$, and the \textbf{Verified 
Schema Knowledge} $\mathcal{K}_t$, which stores only explicitly 
verified metadata and initializes as $\mathcal{K}_0 = \emptyset$.

\vspace{1mm}
\noindent\textbf{Action Space.}
To prevent hallucination, the agent selects actions $a_t \in 
\mathcal{A}$ from four strict categories based on its current 
context $c_t$. The \textit{Explore} action queries database 
metadata. The \textit{Propose} action serves as a mandatory 
cognitive checkpoint at step $t_{\text{propose}}$ to commit 
to the verified schema $\mathcal{K}_{t_{\text{propose}}}$. 
The \textit{Generate} action produces a candidate SQL grounded 
in $\mathcal{K}_t$, and the \textit{Confirm} action submits 
the final SQL query $y$ at the terminal step $T$. 

\vspace{1mm}
\noindent\textbf{Transition and Objective.}
Upon executing $a_t$, the environment emits observation $o_t$ 
and the agent updates its context state to $c_{t+1}$. A 
complete interaction sequence from the agent's perspective 
is represented as a trajectory $\tau = \{(c_t, a_t, o_t)\}_{t=0}^{T}$. 
The ultimate goal of the policy $\pi_\theta(a_t \mid c_t)$ is 
to maximize the expected cumulative reward $J(\theta) = 
\mathbb{E}_{\tau \sim \pi_\theta} [ \sum_{t=0}^{T} \gamma^t r(c_t, a_t) ]$.

\subsection{Reward Components}
\label{subsec:reward}

To evaluate the trajectory, we define three distinct reward 
signals. The specific mechanism for assigning these signals 
to individual tokens is detailed in Section~\ref{subsec:rl_strategy}.

\noindent\textbf{Execution Reward~}($R_{\text{exec}}$).
This reward compares the result $\texttt{Exec}(y)$ of the
predicted SQL against the ground truth
$\texttt{Exec}(y^*)$, with $\bot$ denoting that a query
fails to execute. The reward is assigned as
\begin{equation}
	R_{\text{exec}}(y, y^*) =
	\begin{cases}
		1.0 & \text{if } \texttt{Exec}(y) = \texttt{Exec}(y^*) \\
		0.2 & \text{if } \texttt{Exec}(y) \neq \bot \\
		0.0 & \text{if } \texttt{Exec}(y) = \bot
	\end{cases}
\end{equation}
where the middle case provides partial credit for SQL that
executes but yields an incorrect result.

\noindent\textbf{Format Reward}~($R_{\text{fmt}}$).
This constitutes a trajectory-level signal requiring 
consistent protocol adherence. The reward is defined as
\begin{equation}
	R_{\text{fmt}}(\tau) = 
	\begin{cases}
		0.1 & \text{if protocol is fully adhered to} \\
		0.0 & \text{otherwise}
	\end{cases}
\end{equation}
Full adherence requires that every action $a_t$ conforms 
to prescribed format, all four action categories in 
$\mathcal{A}$ appear at least once, 
and no execution errors occur in the observations $o_t$.

\noindent\textbf{Schema Reward}~($R_{\text{schema}}$).
This reward evaluates the quality of the schema exploration 
phase. It is computed as
\begin{equation}
	R_{\text{schema}}(\hat{\mathcal{K}}, \mathcal{K}^*) = 
	f_{\text{match}}(\hat{\mathcal{K}}, \mathcal{K}^*)
\end{equation}
where $\hat{\mathcal{K}}$ represents the schema proposed by
the agent at step $t_{\text{propose}}$, and $\mathcal{K}^*$
represents the minimal ground truth schema extracted from
$y^*$. The function $f_{\text{match}}$
evaluates their structural overlap. We adopt the sparse
execution-coupled form by default; see
Section~\ref{subsec:schema_reward_design} for the full
definition of $f_{\text{match}}$ and its variants. 
\subsection{Resolving Credit Assignment via Phase-Aware GRPO}
\label{subsec:rl_strategy}

Standard RL combines exploration and generation under a
single reward, making it hard to attribute success or
failure to specific actions in long trajectories. We thus
leverage the structural boundary of the \textit{Propose}
checkpoint to introduce \textbf{Phase-Aware GRPO}, extending
Group Relative Policy Optimization to clearly
separate the learning signals for schema grounding and SQL
generation. In this work, we instantiate Phase-Aware GRPO
with a two-track decomposition (Schema Track and Full
Track); the same formulation generalizes to any number of
structural checkpoints in longer agentic workflows. A
step-by-step contrast against standard GRPO is provided in
Appendix~\ref{app:grpo_contrast}.

\vspace{1mm}
\noindent\textbf{Track Formulation and Rewards.}
For each question $q$, we sample $G$ trajectories and
divide each $\tau^i$ into two optimization tracks
$k \in \{\text{schema}, \text{full}\}$, where the Schema
Track ends at $T_{\text{schema}} = t_{\text{propose}}$ and
the Full Track spans the entire interaction up to
$T_{\text{full}} = T$. Here $t_{\text{propose}}$ is
defined per rollout as the step of the \textbf{latest}
\textit{Propose} action in the trajectory
(Appendix~\ref{app:t_propose}). A dedicated reward $R_k^i$ is 
assigned to each track
\begin{equation}
    R_k^i = 
    \begin{cases}
        R_{\text{schema}}(\hat{\mathcal{K}}_i, \mathcal{K}^*) 
        & \text{if } k = \text{schema} \\
        R_{\text{exec}}(y_i, y^*) + R_{\text{fmt}}(\tau^i) 
        & \text{if } k = \text{full}
    \end{cases}
\end{equation}
ensuring an independent optimization signal for exploration 
quality regardless of generation errors.

\vspace{1mm}
\noindent\textbf{Masked Advantage Computation.}
Advantages are computed via group-relative normalization 
within each track
\begin{equation}
    A_k^i = \frac{R_k^i - \mu_k}{\sigma_k + \epsilon}
\end{equation}
where $\mu_k$ and $\sigma_k$ are the mean and standard 
deviation of the group rewards. We apply strict token-level 
masking where the advantage $A_k^i$ is broadcast exclusively 
to tokens generated within the active steps $t \in [0, T_k]$. 
This is strictly finer-grained than trajectory-level 
weighting, as it prevents exploration rewards from 
incorrectly crediting generation tokens and vice versa. 
Consequently, tokens generated after the \textit{Propose} 
checkpoint receive zero schema advantage.

\vspace{1mm}
\noindent\textbf{Phase-Aware Loss Function.}
Let $\mathcal{L}_k(\theta)$ denote the GRPO loss computed 
over the active tokens for track $k$ using the masked 
advantage $A_k^i$. The total objective combines both tracks
\begin{equation}
    \mathcal{L}(\theta) = 
    \mathcal{L}_{\text{full}}(\theta) +
    \lambda \cdot \mathcal{L}_{\text{schema}}(\theta)
\end{equation}
where $\lambda$ controls the relative contribution of the 
Schema Track. 
By unifying these components, Phase-Aware
GRPO co-optimizes schema grounding and SQL
generation without mixing their learning signals.

\section{Experiments}
\label{sec:experiments}
\subsection{Experimental Setup}
\label{subsec:setup}

\begin{table*}[t]
\centering
\footnotesize
\caption{Execution Accuracy (EX\%) across multiple
benchmarks. \textbf{Gre} denotes single-sample
performance; \textbf{Maj} denotes majority voting.
\textbf{Bold} indicates the best result and
\underline{underline} indicates the second best
within each group.}
\label{tab:main_results}
\setlength{\tabcolsep}{4pt}
\renewcommand{\arraystretch}{1.2}
\begin{tabular}{l c c c c c c c c c c c}
\toprule
\multirow{2}{*}{Method} &
\multirow{2}{*}{\makecell{Schema\\Prefill}} &
\multicolumn{2}{c}{BIRD (dev)} &
\multicolumn{2}{c}{Spider (test)} &
\multicolumn{2}{c}{Spider-DK} &
\multicolumn{2}{c}{Spider-Syn} &
\multicolumn{2}{c}{Spider-Realistic} \\
\cmidrule(lr){3-4} \cmidrule(lr){5-6} \cmidrule(lr){7-8}
\cmidrule(lr){9-10} \cmidrule(lr){11-12}
& & Gre & Maj & Gre & Maj
  & Gre & Maj & Gre & Maj & Gre & Maj \\
\midrule

\rowcolor[HTML]{EBF5FF}
\multicolumn{12}{l}{\textit{~~~3B -- 4B Models}} \\
SQL-R1-3B
    & $\checkmark$
    & -- & 54.6 & -- & 78.9
    & -- & \underline{70.5} & -- & 66.4 & -- & \underline{71.5} \\
SQL-Trail-3B
    & $\checkmark$
    & 50.1 & 55.1 & 77.7 & 84.3
    & -- & -- & -- & -- & -- & -- \\
MTIR-SQL-4B
    & $\checkmark$
    & \underline{63.1} & \underline{64.4} 
    & \underline{83.4} & --
    & \underline{71.2} & -- 
    & \textbf{78.6} & -- 
    & \underline{78.7} & -- \\
\textbf{TRUST-SQL-4B}
    & $\times$
    & \textbf{64.9} & \textbf{67.2}
    & \textbf{82.8} & \textbf{85.0}
    & \textbf{71.6} & \textbf{73.8}
    & \underline{74.7} & \textbf{77.3}
    & \textbf{79.9} & \textbf{82.5} \\
\midrule

\rowcolor[HTML]{FFF3E0}
\multicolumn{12}{l}{\textit{~~~7B -- 8B Models}} \\
OmniSQL-7B
    & $\checkmark$
    & \underline{63.9} & 66.1 
    & \textbf{87.9} & \textbf{88.9}
    & \underline{76.1} & \underline{77.8} 
    & 69.7 & 69.6 
    & 76.2 & 78.0 \\
SQL-R1-7B
    & $\checkmark$
    & 63.7 & \underline{66.6} 
    & -- & \underline{88.7}
    & -- & \textbf{78.1} 
    & -- & 76.7 
    & -- & 83.3 \\
SQL-Trail-7B
    & $\checkmark$
    & 60.1 & 64.2 
    & \underline{86.0} & 87.6
    & \textbf{76.8} & \underline{77.8}
    & 72.8 & \underline{77.0} 
    & \underline{79.6} & \underline{83.9} \\
MTIR-SQL-8B
    & $\checkmark$
    & -- & 64.6 
    & 83.4 & --
    & 72.9 & -- 
    & \underline{77.2} & -- 
    & 77.4 & -- \\
\textbf{TRUST-SQL-8B}
    & $\times$
    & \textbf{65.8} & \textbf{67.7}
    & 83.9 & 86.5
    & 72.1 & 75.7
    & \textbf{75.4} & \textbf{77.4}
    & \textbf{82.1} & \textbf{84.1} \\
\midrule

\rowcolor[HTML]{F3E5F5}
\multicolumn{12}{l}{\textit{~~~14B Models}} \\
OmniSQL-14B
    & $\checkmark$
    & 64.2 & 65.9
    & \textbf{88.3} & \underline{88.3}
    & 72.9 & 74.8
    & 69.0 & 72.0
    & 76.4 & 78.5 \\
SQL-R1-14B
    & $\checkmark$
    & -- & 67.1
    & -- & 88.1
    & -- & \textbf{79.3}
    & -- & \underline{78.5}
    & -- & \underline{86.2} \\
SQL-Trail-14B
    & $\checkmark$
    & 63.6 & 66.7
    & 86.8 & \textbf{88.5}
    & -- & --
    & -- & --
    & -- & -- \\
MTIR-SQL-14B
    & $\checkmark$
    & \underline{67.2} & \underline{68.1}
    & \underline{87.2} & --
    & \textbf{76.3} & --
    & \textbf{81.0} & --
    & \underline{81.1} & -- \\
\textbf{TRUST-SQL-14B}
    & $\times$
    & \textbf{68.1} & \textbf{70.1}
    & 83.9 & 86.5
    & \underline{75.0} & \underline{78.6}
    & \underline{78.4} & \textbf{79.7}
    & \textbf{82.7} & \textbf{86.5} \\
\bottomrule
\end{tabular}
\end{table*}

\noindent\textbf{Implementation Details.}
We use Qwen3-4B, Qwen3-8B, and Qwen3-14B as our base
models and implement all experiments using the SLIME
framework~\citep{slime}. Each model is trained in two
sequential stages: SFT warm-up followed by Phase-Aware
GRPO optimization.
An ablation on the necessity of the SFT warm-up stage is
provided in Section~\ref{subsec:cold_start}. Further
details are provided in Appendix~\ref{app:implementation}.

\noindent\textbf{Baselines.}
TRUST-SQL utilizes a highly efficient data recipe comprising 
9.2k SFT samples and 11.6k RL samples. We compare our 
framework against recent strong baselines across the 3B to 14B
parameter scales. Single-turn models include OmniSQL~\citep{omnisql} 
and SQL-R1~\citep{sql_r1}. Multi-turn RL methods include 
MTIR-SQL~\citep{mtir} and SQL-Trail~\citep{sqltrail}. Full 
dataset construction and detailed baseline comparisons are 
provided in Appendix~\ref{app:data}.

\noindent\textbf{Evaluation Benchmarks and Metrics.}
We evaluate on BIRD-Dev~\citep{bird} for large-scale schema 
grounding and Spider-Test~\citep{spider} for compositional 
generalization. To stress-test model robustness, we incorporate 
three challenging variants. Specifically, Spider-Syn~\citep{spider_syn} 
evaluates lexical robustness via synonym substitution, 
Spider-DK~\citep{spider_dk} probes for implicit domain knowledge, 
and Spider-Realistic~\citep{spider_real} assesses ambiguity 
resolution. We measure Execution Accuracy where the predicted 
SQL must yield the exact same database result as the ground truth. 
We report single-sample performance via Greedy decoding at 
temperature zero and execution-based Majority voting across 
multiple sampled queries.

\subsection{Main Results}
\label{subsec:main_results}

Table~\ref{tab:main_results} presents the execution accuracy
across all benchmarks. For the majority voting evaluation,
we sample trajectories at a temperature of 0.8 with a 15-turn
inference budget, as analyzed in
Section~\ref{subsec:test_time_scaling}.
Detailed token consumption and tool
invocation statistics are provided in Appendix~\ref{app:cost_analysis}.

\noindent\textbf{Scaling Across 4B / 8B / 14B.}
TRUST-SQL performs competitively at all three scales while
operating under the Unknown Schema setting. On BIRD-Dev
specifically,
the 4B / 8B / 14B variants reach 64.9 / 65.8 / 68.1 with
greedy decoding and 67.2 / 67.7 / 70.1 with majority
voting. TRUST-SQL-14B outperforms every 14B baseline that
uses full schema prefilling on this benchmark. On the
robustness suite,
TRUST-SQL leads Spider-Realistic at all three scales
(82.5 / 84.1 / 86.5 majority) and Spider-Syn majority
voting at all three scales (77.3 / 77.4 / 79.7); on
Spider-DK, the 4B variant tops its group (71.6 / 73.8)
while the 8B and 14B variants remain within a few points
of the schema-prefilled best. On the standard Spider-Test
set, where schema-prefilled baselines benefit from
explicit mapping cues, TRUST-SQL remains within a few
points without any pre-loaded schema. These results
suggest that the four-phase protocol and Phase-Aware GRPO
continue to provide gains as scale grows.

\noindent\textbf{The Value of Autonomous Exploration.}
TRUST-SQL achieves these scores under the strict Unknown
Schema setting, while all baselines rely on full schema
prefilling that consumes substantial context and assumes
perfect database observability. Further comparisons
against strong proprietary agentic models and a serving-cost analysis are
reported in Appendices~\ref{app:large_agentic}
and~\ref{app:cost_pareto}.

\subsection{Can Schema Prefill Boost Performance?}
\label{subsec:schema_prefill}
While TRUST-SQL operates without any pre-loaded schema, 
a natural question arises as to whether injecting the 
complete schema would further boost performance. We thus 
introduce a Schema Prefill variant where the full schema 
is delivered as a single synthetic \textit{Explore} turn 
at the beginning of the interaction, providing all table 
and column information at once. The case study is shown 
in Appendix~\ref{app:case_study}.

\newcommand{\up}[1]{\textcolor[HTML]{D32F2F}{\scriptsize~$\uparrow$#1}}
\newcommand{\down}[1]{\textcolor[HTML]{2E7D32}{\scriptsize~$\downarrow$#1}}

\begin{table}[htbp]
\centering
\caption{Effect of Schema Prefill (greedy decoding). Arrows 
denote absolute accuracy changes compared to the Unknown 
Schema ($\times$) setting.}
\label{tab:schema_prefill}
\setlength{\tabcolsep}{2pt}
\renewcommand{\arraystretch}{1.0}
\footnotesize 
\begin{tabular}{c ccccc}
\toprule
Prefill & BIRD & Spider & S-DK & S-Syn & S-Realistic \\
\midrule

\rowcolor[HTML]{EBF5FF}
\multicolumn{6}{l}{\textbf{Qwen3-4B}} \\
$\times$
    & 29.3 & 51.2 & 43.7 & 47.4 & 49.2 \\
$\checkmark$
    & 46.3\up{17.0} & 67.6\up{16.4} & 57.0\up{13.3} & 62.6\up{15.2} & 65.9\up{16.7} \\
\rowcolor[HTML]{EBF5FF}
\multicolumn{6}{l}{\textbf{TRUST-SQL-4B}} \\
$\times$
    & 64.9 & 82.8 & 71.6 & 74.7 & 79.9 \\
$\checkmark$
    & 64.8\down{0.1} & 83.1\up{0.3} & 69.2\down{2.4} & 72.5\down{2.2} & 80.1\up{0.2} \\
\midrule

\rowcolor[HTML]{FFF3E0}
\multicolumn{6}{l}{\textbf{Qwen3-8B}} \\
$\times$
    & 47.9 & 67.4 & 56.3 & 58.4 & 66.3 \\
$\checkmark$
    & 49.9\up{2.0} & 68.3\up{0.9} & 57.6\up{1.3} & 64.5\up{6.1} & 68.1\up{1.8} \\
\rowcolor[HTML]{FFF3E0}
\multicolumn{6}{l}{\textbf{TRUST-SQL-8B}} \\
$\times$
    & 65.8 & 83.9 & 72.1 & 75.4 & 82.1 \\
$\checkmark$
    & 65.5\down{0.3} & 84.0\up{0.1} & 74.4\up{2.3} & 75.4 & 80.5\down{1.6} \\
\bottomrule
\end{tabular}
\vspace{-2mm}
\end{table}

As shown in Table~\ref{tab:schema_prefill}, the base Qwen3
models rely heavily on pre-loaded metadata.
Without schema prefilling, their performance drops sharply, evidenced by
a 17.0\% absolute drop for Qwen3-4B on BIRD. This
suggests that standard models do not exhibit autonomous exploration capabilities. When equipped with our framework, TRUST-SQL closes this gap, with substantial gains
over the base models. For
instance, TRUST-SQL-4B yields a \textbf{35.6\% absolute
improvement} over Qwen3-4B on BIRD.
Across all five benchmarks,
the framework delivers an average absolute improvement of
\textbf{30.6\%} for the 4B variant and \textbf{16.6\%} for the 8B variant
compared to their respective base models under the Unknown Schema setting.

Furthermore, TRUST-SQL is largely insensitive to
pre-loaded schemas. For both 4B and 8B variants,
injecting the full schema upfront provides only negligible
changes on BIRD and Spider. In some cases it slightly degrades
performance on robustness benchmarks. Specifically, TRUST-SQL-4B drops by 2.4\% on Spider-DK and
TRUST-SQL-8B drops by 1.6\% on Spider-Realistic. The iterative policy
already retrieves the metadata necessary for SQL generation,
making full schema injection largely redundant.
Therefore, \textbf{active exploration is a viable
alternative} to static prefilling.

\section{Analysis}
\label{sec:analysis}
\subsection{What Makes a Good Schema Reward?}
\label{subsec:schema_reward_design}

We investigate two key design dimensions for
$f_{\text{match}}$ defined in Section~\ref{subsec:reward}:
whether $R_{\text{schema}}$ should be \textit{coupled}
with $R_{\text{exec}}$, and whether $f_{\text{match}}$
should be \textit{sparse} or \textit{dense}.
Specifically,
\textbf{Sparse + Uncoupled} assigns $R_{\text{schema}}$
regardless of $R_{\text{exec}}$ with a binary $f_{\text{match}}$.
\textbf{Sparse + Coupled} (\textbf{TRUST-SQL}) conditions
$R_{\text{schema}}$ on $R_{\text{exec}} = 1.0$ with the
same binary criterion. \textbf{Dense + Coupled} further
replaces $f_{\text{match}}$ with a graduated function
enforcing full recall as a hard gate before assigning
partial precision-based rewards. Formal definitions of all
three variants follow.

Let $P_t, P_c$ denote the proposed table and column sets
in $\hat{\mathcal{K}}$, and $G_t, G_c$ the corresponding
ground-truth sets in $\mathcal{K}^*$. For the dense variant,
$r_t$ and $r_c$ denote table and column recall, while
$p_t$ and $p_c$ denote the corresponding precision values;
$w_t$ and $w_c$ are their weighting coefficients.

\noindent\textbf{(i) Sparse + Uncoupled.}
Exact-match reward on both tables and columns:
\begin{equation}
    f_{\text{match}}
    = \mathbf{1}\!\left[P_t = G_t \,\land\, P_c = G_c\right].
\end{equation}

\noindent\textbf{(ii) Sparse + Coupled (default).}
The same exact-match criterion, additionally gated by
execution correctness so that schema credit is given only
when the resulting SQL also executes correctly:
\begin{equation}
    R_{\text{schema}}
    = \mathbf{1}[R_{\text{exec}}=1] \cdot f_{\text{match}}.
\end{equation}

\noindent\textbf{(iii) Dense + Coupled.}
First enforce full recall on both tables and columns
($r_t = r_c = 1$); then compute a discretized reward based
on a weighted combined precision
\begin{equation}
    \pi = w_t\, p_t + w_c\, p_c,
\end{equation}
with reward levels $\{1.0,0.8,0.6,0.4,0.2,0.1\}$ at
thresholds $\{1.0,0.9,0.8,0.7,0.6\}$, again gated by
execution success. Sparse variants reward only exact schema
identification, whereas the dense coupled variant strongly
penalizes missing schema elements while still assigning
partial credit to over-selected columns, provided that the
final SQL executes correctly.
\begin{figure}[htbp]
	\centering
	\includegraphics[width=0.9\columnwidth]{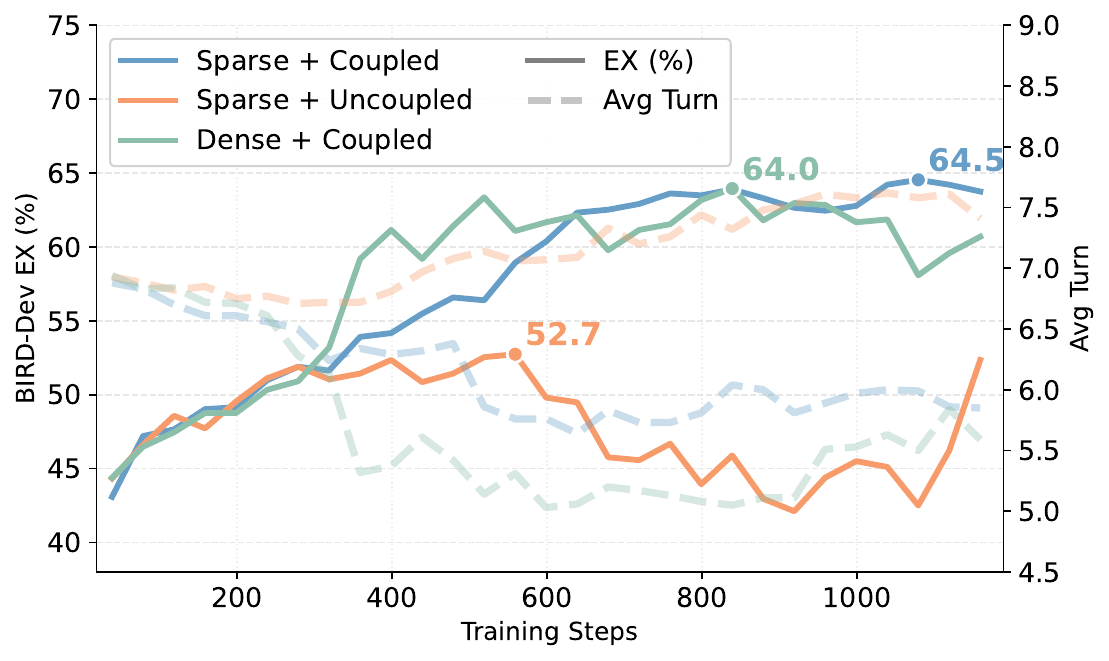}
	\caption{Ablation on schema reward formulation.}
	\label{fig:reward_design}
	\vspace{-2mm}
\end{figure}

As shown in Figure~\ref{fig:reward_design}, the three 
variants exhibit markedly different training dynamics. 
\textbf{Sparse + Uncoupled} achieves the lowest EX of 
52.7\% despite the highest turn count of 6.71, revealing 
that decoupling schema reward from execution incentivizes 
redundant exploration rather than precise grounding. 
\textbf{Dense + Coupled} reduces turns to 5.03 but 
converges to a suboptimal 64.0\%, as the graduated 
$f_{\text{match}}$ introduces conflicting gradients between 
maximizing recall and minimizing unnecessary columns. 
\textbf{Sparse + Coupled} achieves the best EX of 64.5\% 
with a balanced turn count of 5.64, where the binary 
$f_{\text{match}}$ provides an unambiguous optimization 
target and conditioning on $R_{\text{exec}} = 1.0$ 
establishes a direct causal chain between exploration 
quality and task success. These results indicate that 
\textit{coupling} is a more critical design dimension
than \textit{reward density}.

\subsection{Is Cold-Start SFT Necessary?}
\label{subsec:cold_start}

TRUST-SQL adopts a two-stage training pipeline in which Phase-Aware GRPO is preceded by an SFT warm-up phase. To assess its necessity, we compare three training configurations on TRUST-SQL-4B (Table~\ref{tab:cold_start}). Applying Phase-Aware GRPO directly without SFT warm-up yields 59.9\% on BIRD and 79.6\% on Spider, both substantially below the full pipeline. These numbers are largely illusory: without SFT initialization, the model quickly learns to \textbf{hack the reward by exhaustively querying all tables and columns in the first turn}, completing the interaction in roughly four actions. This degenerates the Unknown Schema setting into a disguised Full Schema scenario, bypassing genuine active exploration. SFT alone achieves reasonable performance, confirming that the warm-up phase instills structured exploration behavior, while the full two-stage pipeline consistently achieves the best results. This indicates that Phase-Aware GRPO provides gains that cannot be attributed to supervised learning alone.

\begin{table}[h]
	\centering
	\small
	\setlength{\tabcolsep}{4pt}
	\renewcommand{\arraystretch}{1.2}
	\caption{Ablation on cold-start SFT for TRUST-SQL-4B
		(greedy decoding).}
	\label{tab:cold_start}
	\begin{tabular}{l cc cc}
		\toprule
		\textbf{Configuration} & \textbf{SFT} & \textbf{RL}
		& \textbf{BIRD (dev)} & \textbf{Spider (test)} \\
		\midrule
		SFT Only
		& \textcolor[HTML]{2E7D32}{\ding{51}}
		& \textcolor[HTML]{D32F2F}{\ding{55}}
		& 46.2 & 66.7 \\
		RL Only
		& \textcolor[HTML]{D32F2F}{\ding{55}}
		& \textcolor[HTML]{2E7D32}{\ding{51}}
		& 59.9 & 79.6 \\
		\rowcolor[HTML]{EBF5FF}
		SFT + RL
		& \textcolor[HTML]{2E7D32}{\ding{51}}
		& \textcolor[HTML]{2E7D32}{\ding{51}}
		& \textbf{64.9} & \textbf{82.8} \\
		\bottomrule
	\end{tabular}
\end{table}

\subsection{Performance on Complex Benchmark (Spider 2.0)}
\label{subsec:spider2}

To evaluate TRUST-SQL under more challenging real-world
conditions, we conduct additional experiments on the
SQLite subset of Spider 2.0~\citep{spider2}, comprising
135 questions over large-scale databases featuring
significantly more complex schemas and larger table counts
than standard Spider. 
Table~\ref{tab:spider2} reports execution accuracy 
alongside representative baselines. Notably, strong 
proprietary models such as GPT-4o~\citep{gpt4o} and 
DeepSeek-V3~\citep{deepseekv3} achieve only 15.6\% on 
this benchmark, while specialized Text-to-SQL models 
like OmniSQL-7B~\citep{omnisql} reach 10.4\%, reflecting 
the substantial difficulty of this setting.

\begin{table}[H]
	\centering
	\small
	\caption{Execution accuracy on the Spider 2.0 SQLite 
		subset (135 questions). Baselines use full schema 
		prefilling. Pass@8 is computed over 8 sampled trajectories.}
	\label{tab:spider2}
	\setlength{\tabcolsep}{4pt}
	\renewcommand{\arraystretch}{1.1}
	\resizebox{\columnwidth}{!}{%
		\begin{tabular}{l c c c}
			\toprule
			\textbf{Method} 
			& \textbf{Prefill} 
			& \textbf{Greedy} 
			& \textbf{Pass@8} \\
			\midrule
			OmniSQL-7B
			& \textcolor[HTML]{2E7D32}{\ding{51}} 
			& 10.4 & -- \\
			GPT-4o
			& \textcolor[HTML]{2E7D32}{\ding{51}} 
			& 15.6 & -- \\
			DeepSeek-V3
			& \textcolor[HTML]{2E7D32}{\ding{51}} 
			& 15.6 & -- \\
			\midrule
			OpenSearchSQL+Qwen2.5-7B-Instruct
			& \textcolor[HTML]{2E7D32}{\ding{51}} 
			& 4.4 & 7.4 \\
			OpenSearchSQL+Arctic-Text2SQL-R1-7B
			& \textcolor[HTML]{2E7D32}{\ding{51}} 
			& 14.1 & 20.7 \\
			\midrule
			\rowcolor[HTML]{FFF3E0}
			\textbf{TRUST-SQL-8B}
			& \textcolor[HTML]{D32F2F}{\ding{55}} 
			& \textbf{14.8} & \textbf{24.9} \\
			\bottomrule
		\end{tabular}%
	}
\end{table}

Despite operating entirely without pre-loaded metadata, 
TRUST-SQL-8B achieves 14.8\% greedy accuracy and 24.9\% 
Pass@8, surpassing OpenSearchSQL paired with the 
specialized Arctic-7B~\citep{arctic} model. 
The non-saturating Pass@8 curve 
further suggests substantial headroom for improvement
with increased sampling budgets, validating the
generalizability of our framework beyond standard
academic benchmarks.

\begin{figure*}[!t]
	\centering
	\begin{subfigure}[t]{0.32\textwidth}
		\includegraphics[width=\linewidth]
		{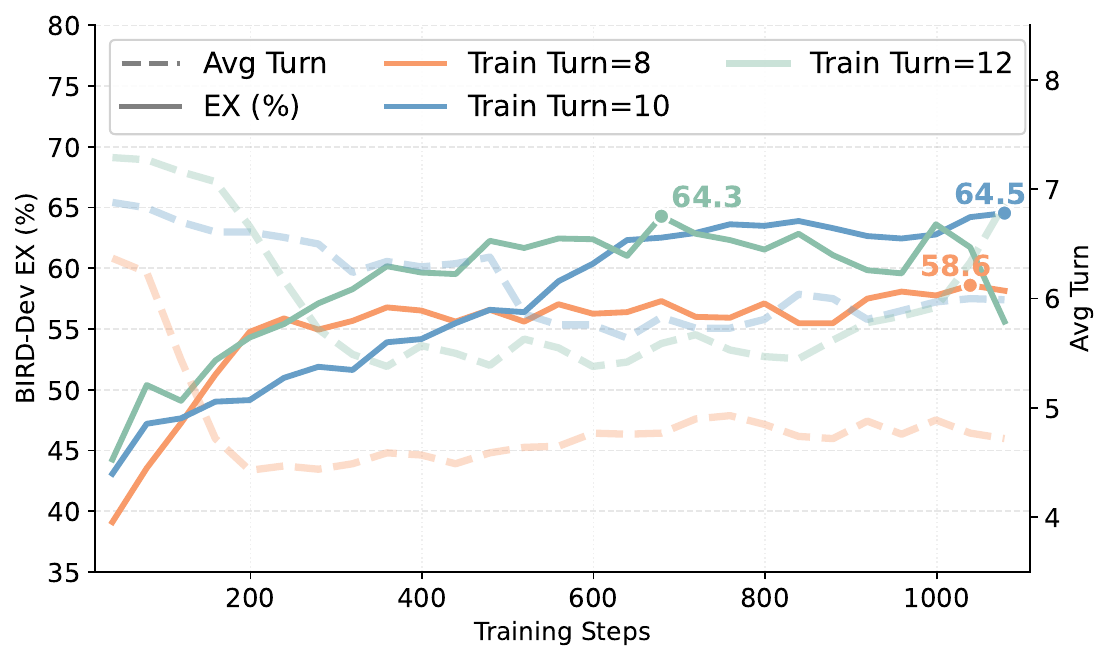}
		\caption{Training turn budget effect.}
		\label{fig:scaling_train_turn}
	\end{subfigure}
	\hfill
	\begin{subfigure}[t]{0.32\textwidth}
		\includegraphics[width=\linewidth]
		{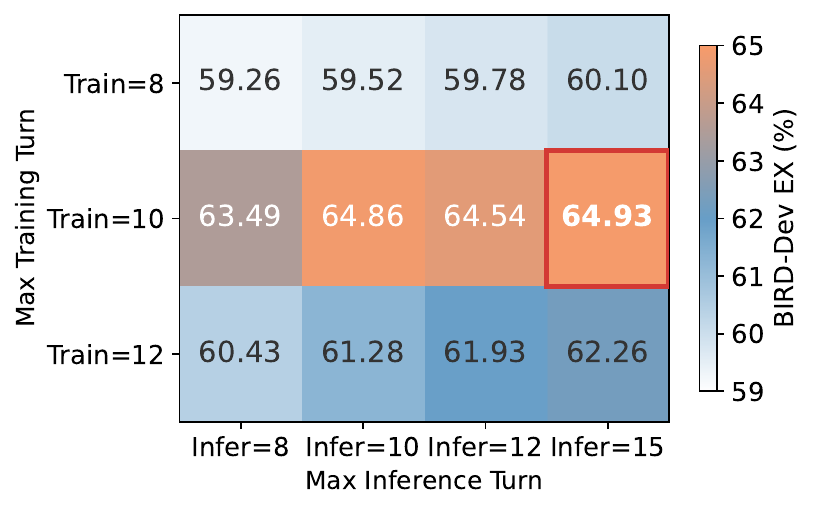}
		\caption{Train vs.\ inference turn budget.}
		\label{fig:scaling_heatmap}
	\end{subfigure}
	\hfill
	\begin{subfigure}[t]{0.32\textwidth}
		\includegraphics[width=\linewidth]
		{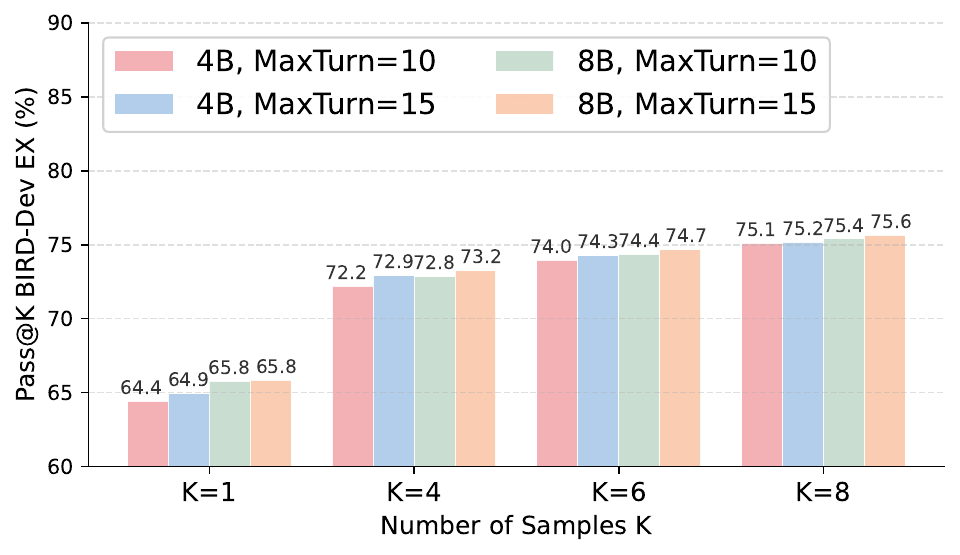}
		\caption{Pass@$K$ scaling.}
		\label{fig:scaling_turn_passK}
	\end{subfigure}
	\caption{Test-time scaling analysis of TRUST-SQL
		across three dimensions: training turn budget,
		training vs.\ inference turn budget interaction,
		and Pass@$K$ with repeated sampling.}
	\label{fig:test_time_scaling}
	\vspace{-2mm}
\end{figure*}
\subsection{Test-Time Scaling Behavior}
\label{subsec:test_time_scaling}

We analyze the test-time scaling properties of TRUST-SQL
across three dimensions.

\noindent\textbf{Training Turn Budget.}
As illustrated in Figure~\ref{fig:scaling_train_turn},
expanding the training turn budget from 8 to 10 yields
substantial gains on BIRD-Dev. However, further increasing
to 12 turns causes severe training instability where the
average turn count spikes and execution accuracy sharply
declines, suggesting that an overly permissive horizon
fails to penalize redundant exploration. Consequently,
a 10-turn budget provides the optimal balance between
accuracy and exploration efficiency.

\noindent\textbf{Interaction Between Horizons.}
As shown in Figure~\ref{fig:scaling_heatmap}, a 10-turn
training budget consistently yields the strongest baseline
policy. Providing additional inference turns beyond the
training horizon further improves performance, with the
optimal configuration pairing a 10-turn training budget
with a 15-turn inference budget to achieve a peak accuracy
of \textbf{64.93\%}. This demonstrates that the agent
effectively utilizes extra test-time compute to recover
from early exploration mistakes.

\noindent\textbf{Scaling with Repeated Sampling.}
As shown in Figure~\ref{fig:scaling_turn_passK}, all
configurations exhibit monotonic accuracy improvements as
the sample size $K$ grows, driven by \textit{exploration
	diversity} across independently sampled trajectories. The
persistent gap between Pass@$K$ and greedy performance
indicates that the model can generate correct solutions
but has not fully converged to a consistent policy,
suggesting headroom for further RL training.

\subsection{How to Balance Exploration and Generation?}
\label{subsec:lambda_ablation}

In the Phase-Aware GRPO loss, $\lambda$ controls the relative contribution of the Schema Track. We ablate $\lambda \in {0.125, 0.25, 0.375}$ against two $\lambda=0$ baselines. $w/o$ $schema$ uses only the execution outcome, while $w/ schema$ adds the schema reward with a fixed weight of 0.25 to the terminal reward without separate tracks.
\begin{figure}[h]
	\centering
	\begin{subfigure}[b]{0.49\columnwidth}
		\centering
		\includegraphics[width=\linewidth]{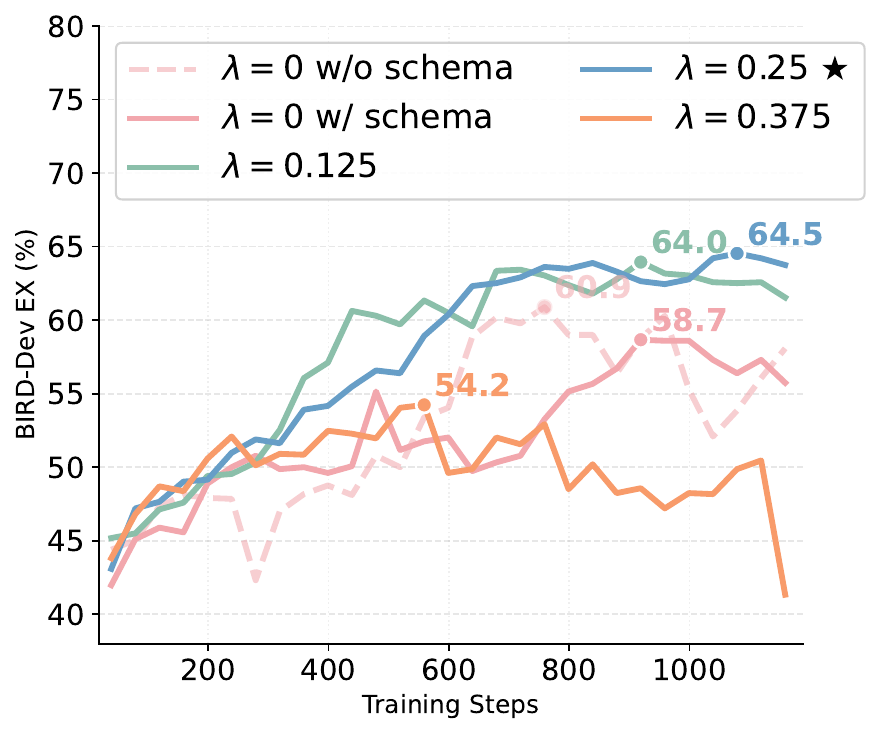}
		\caption{EX (\%) during training.}
		\label{fig:lambda_ablation_ex}
	\end{subfigure}
	\hfill
	\begin{subfigure}[b]{0.49\columnwidth}
		\centering
		\includegraphics[width=\linewidth]{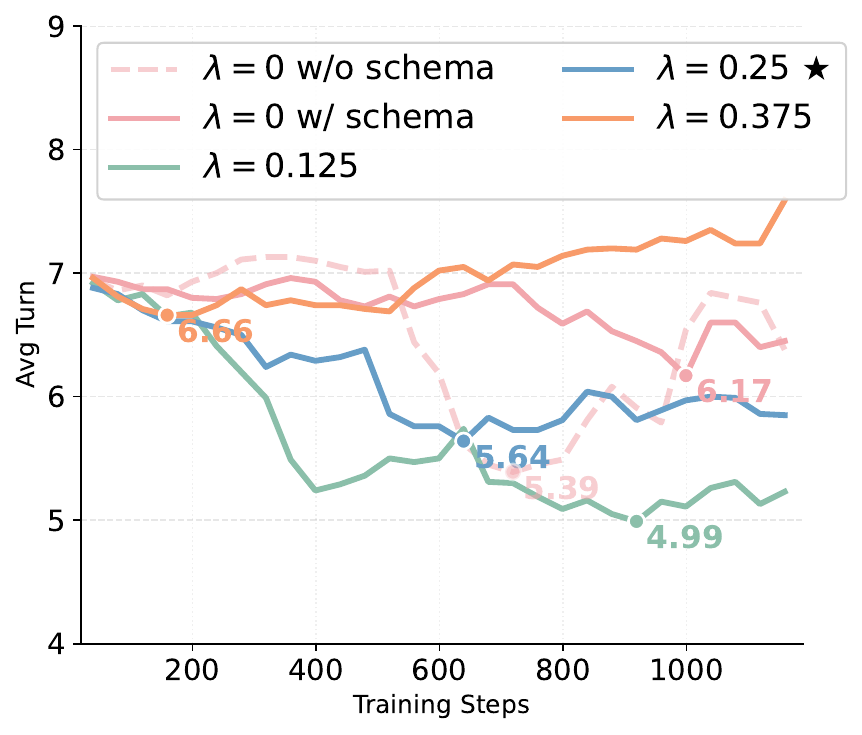}
		\caption{Avg turns during training.}
		\label{fig:lambda_ablation_turn}
	\end{subfigure}
	\caption{Effect of $\lambda$ on training dynamics. 
	}
	\label{fig:lambda_ablation}
\end{figure}

As shown in Figure~\ref{fig:lambda_ablation_ex}, naively 
mixing the schema reward into the terminal step yields 58.7\%, 
worse than the 60.9\% achieved by the pure execution baseline. 
This confirms that conflating exploration and generation 
obscures the reward signal. In contrast, the optimal Phase-Aware
setting at $\lambda = 0.25$ peaks at \textbf{64.5\%}, yielding 
a \textbf{+5.8\%} gain over naive aggregation and a 
\textbf{+3.6\%} gain over the pure execution baseline.
Furthermore, $\lambda$ dictates the balance between exploration 
and generation. While $\lambda = 0.125$ achieves a competitive 
64.0\%, an excessively large $\lambda = 0.375$ severely degrades 
performance to 54.2\%. As shown in Figure~\ref{fig:lambda_ablation_turn}, 
this over-weighted schema reward incentivizes the agent to 
remain perpetually in the exploration phase, causing a sharp 
increase in average interaction turns and over-optimizing 
metadata retrieval at the expense of SQL generation.

 \section{Conclusion}
\label{sec:conclusion}
\vspace{-1mm}
In this work, we revisit the \textbf{Full Schema Assumption}
that underlies Text-to-SQL research.
By formalizing the task as a POMDP under the \textbf{Unknown Schema} setting,
\textbf{TRUST-SQL} demonstrates that autonomous database
exploration is both feasible and effective under partial
schema observability, where the full database structure
cannot be assumed to be available.
The structured \textbf{four-phase protocol}
grounds agent reasoning in actively verified metadata to
suppress hallucinations, while its mandatory cognitive
checkpoint provides a structural boundary for \textbf{Phase-Aware
GRPO} to resolve the credit assignment bottleneck, yielding
a \textbf{9.9\%} relative improvement over standard GRPO.
Experiments across five benchmarks at three scales confirm the framework's effectiveness.
Despite operating without pre-loaded metadata, TRUST-SQL
matches or surpasses schema-prefilled baselines on most
benchmarks, advancing reliable Text-to-SQL in unobservable environments.
\newpage
\section*{Limitations}
\label{sec:limitations}

While TRUST-SQL demonstrates strong performance under the 
Unknown Schema setting, several limitations remain. 

\noindent\textbf{Inference Overhead.}
The multi-turn interaction paradigm naturally incurs higher 
inference cost compared to single-turn methods, as each 
interaction step involves a live database call. However, 
as shown in Appendix~\ref{app:cost_analysis}, this overhead 
remains modest in practice. 
Further optimizing inference efficiency for 
latency-critical deployments remains a practical direction 
for future work.


\noindent\textbf{SQLite Dialect Only.}
Both training and evaluation are conducted on SQLite-based 
benchmarks, as BIRD and Spider exclusively use SQLite. 
Extending to other SQL dialects such as PostgreSQL or 
MySQL remains a valuable direction for future work.

\noindent\textbf{Fixed Turn Budget.}
The maximum interaction turn $T$ is fixed at training 
time, which may limit exploration thoroughness for 
databases with exceptionally complex schemas. Adapting 
the turn budget dynamically based on database complexity 
remains an interesting direction for future work.

\section*{Reproducibility Statement}

To ensure full reproducibility, we release the complete 
source code at \url{https://github.com/JaneEyre0530/TrustSQL}. 
All dataset construction pipelines are detailed in 
Appendix~\ref{app:data}, training hyperparameters and 
hardware specifications are summarized in 
	Appendix~\ref{app:implementation}.

\bibliography{custom}

@misc{autolink,
	title={AutoLink: Autonomous Schema Exploration and Expansion for Scalable Schema Linking in Text-to-SQL at Scale}, 
	author={Ziyang Wang and Yuanlei Zheng and Zhenbiao Cao and Xiaojin Zhang and Zhongyu Wei and Pei Fu and Zhenbo Luo and Wei Chen and Xiang Bai},
	year={2025},
	eprint={2511.17190},
	archivePrefix={arXiv},
	primaryClass={cs.CL},
	url={https://arxiv.org/abs/2511.17190}, 
}

@article{route,
	title={ROUTE: Robust multitask tuning and collaboration for Text-to-SQL},
	author={Qin, Yang and Chen, Chao and Fu, Zhihang and Chen, Ze and Peng, Dezhong and Hu, Peng and Ye, Jieping},
	journal={arXiv preprint arXiv:2412.10138},
	year={2024}
}

@inproceedings{star,
	title={Star-sql: Self-taught reasoner for text-to-sql},
	author={He, Mingqian and Shen, Yongliang and Zhang, Wenqi and Peng, Qiuying and Wang, Jun and Lu, Weiming},
	booktitle={Proceedings of the 63rd Annual Meeting of the Association for Computational Linguistics (Volume 1: Long Papers)},
	pages={24365--24375},
	year={2025}
}

@article{sqltrail,
	title={SQL-Trail: Multi-Turn Reinforcement Learning with Interleaved Feedback for Text-to-SQL},
	author={Hua, Harper and Han, Zhen and Shen, Zhengyuan and Lee, Jeremy and Guan, Patrick and Zhu, Qi and Jeoung, Sullam and Chen, Yueyan and Bai, Yunfei and Wang, Shuai and others},
	journal={arXiv preprint arXiv:2601.17699},
	year={2026}
}

@article{omnisql,
	title={Omnisql: Synthesizing high-quality text-to-sql data at scale},
	author={Li, Haoyang and Wu, Shang and Zhang, Xiaokang and Huang, Xinmei and Zhang, Jing and Jiang, Fuxin and Wang, Shuai and Zhang, Tieying and Chen, Jianjun and Shi, Rui and others},
	journal={arXiv preprint arXiv:2503.02240},
	year={2025}
}

@article{sql_r1,
	title={Sql-r1: Training natural language to sql reasoning model by reinforcement learning},
	author={Ma, Peixian and Zhuang, Xialie and Xu, Chengjin and Jiang, Xuhui and Chen, Ran and Guo, Jian},
	journal={arXiv preprint arXiv:2504.08600},
	year={2025}
}

@article{reward_sql,
	title={Reward-sql: Boosting text-to-sql via stepwise reasoning and process-supervised rewards},
	author={Zhang, Yuxin and Fan, Meihao and Fan, Ju and Yi, Mingyang and Luo, Yuyu and Tan, Jian and Li, Guoliang},
	journal={arXiv preprint arXiv:2505.04671},
	year={2025}
}

@article{reasoning_sql,
	title={Reasoning-sql: Reinforcement learning with sql tailored partial rewards for reasoning-enhanced text-to-sql},
	author={Pourreza, Mohammadreza and Talaei, Shayan and Sun, Ruoxi and Wan, Xingchen and Li, Hailong and Mirhoseini, Azalia and Saberi, Amin and Arik, Sercan and others},
	journal={arXiv preprint arXiv:2503.23157},
	year={2025}
}

@article{mtir,
	title={MTIR-SQL: Multi-turn Tool-Integrated Reasoning Reinforcement Learning for Text-to-SQL},
	author={Xu, Zekun and Xia, Siyu and Yue, Chuhuai and Chai, Jiajun and Tian, Mingxue and Wang, Xiaohan and Lin, Wei and Li, Haoxuan and Yin, Guojun},
	journal={arXiv preprint arXiv:2510.25510},
	year={2025}
}

@article{arctic,
	title={Arctic-text2sql-r1: Simple rewards, strong reasoning in text-to-sql},
	author={Yao, Zhewei and Sun, Guoheng and Borchmann, Lukasz and Nuti, Gaurav and Shen, Zheyu and Deng, Minghang and Zhai, Bohan and Zhang, Hao and Li, Ang and He, Yuxiong},
	journal={arXiv preprint arXiv:2505.20315},
	year={2025}
}

@misc{duca,
	title={Harmonizing Dense and Sparse Signals in Multi-turn RL: Dual-Horizon Credit Assignment for Industrial Sales Agents}, 
	author={Haojin Yang and Ai Jian and Xinyue Huang and Yiwei Wang and Weipeng Zhang and Ke Zeng and Xunliang Cai and Jingqing Ruan},
	year={2026},
	eprint={2603.01481},
	archivePrefix={arXiv},
	primaryClass={cs.AI},
	url={https://arxiv.org/abs/2603.01481}, 
}

@article{lost,
	title={Llms get lost in multi-turn conversation},
	author={Laban, Philippe and Hayashi, Hiroaki and Zhou, Yingbo and Neville, Jennifer},
	journal={arXiv preprint arXiv:2505.06120},
	year={2025}
}

@article{sweet_rl,
	title={Sweet-rl: Training multi-turn llm agents on collaborative reasoning tasks},
	author={Zhou, Yifei and Jiang, Song and Tian, Yuandong and Weston, Jason and Levine, Sergey and Sukhbaatar, Sainbayar and Li, Xian},
	journal={arXiv preprint arXiv:2503.15478},
	year={2025}
}

@article{spider2,
	title={Spider 2.0: Evaluating language models on real-world enterprise text-to-sql workflows},
	author={Lei, Fangyu and Chen, Jixuan and Ye, Yuxiao and Cao, Ruisheng and Shin, Dongchan and Su, Hongjin and Suo, Zhaoqing and Gao, Hongcheng and Hu, Wenjing and Yin, Pengcheng and others},
	journal={arXiv preprint arXiv:2411.07763},
	year={2024}
}

@inproceedings{spider,
    title     = {Spider: A Large-Scale Human-Labeled Dataset for Complex and Cross-Domain Semantic Parsing and Text-to-SQL Task},
    author    = {Tao Yu and Rui Zhang and Kai Yang and Michihiro Yasunaga and Dongxu Wang and Zifan Li and James Ma and Irene Li and Qingning Yao and Shanelle Roman and Zilin Zhang and Dragomir Radev},
    booktitle = {Proceedings of the 2018 Conference on Empirical Methods in Natural Language Processing},
    address   = {Brussels, Belgium},
    publisher = {Association for Computational Linguistics},
    year      = {2018}
}

@article{bird,
	title={Can llm already serve as a database interface? a big bench for large-scale database grounded text-to-sqls},
	author={Li, Jinyang and Hui, Binyuan and Qu, Ge and Yang, Jiaxi and Li, Binhua and Li, Bowen and Wang, Bailin and Qin, Bowen and Geng, Ruiying and Huo, Nan and others},
	journal={Advances in Neural Information Processing Systems},
	volume={36},
	year={2024}
}

@misc{scale,
	title={Agentar-Scale-SQL: Advancing Text-to-SQL through Orchestrated Test-Time Scaling}, 
	author={Pengfei Wang and Baolin Sun and Xuemei Dong and Yaxun Dai and Hongwei Yuan and Mengdie Chu and Yingqi Gao and Xiang Qi and Peng Zhang and Ying Yan},
	year={2025},
	eprint={2509.24403},
	archivePrefix={arXiv},
	primaryClass={cs.CL},
	url={https://arxiv.org/abs/2509.24403}, 
}

@misc{metadata,
	title={Automatic Metadata Extraction for Text-to-SQL}, 
	author={Vladislav Shkapenyuk and Divesh Srivastava and Theodore Johnson and Parisa Ghane},
	year={2025},
	eprint={2505.19988},
	archivePrefix={arXiv},
	primaryClass={cs.DB},
	url={https://arxiv.org/abs/2505.19988}, 
}

@misc{ragen,
	title={RAGEN: Understanding Self-Evolution in LLM Agents via Multi-Turn Reinforcement Learning}, 
	author={Zihan Wang and Kangrui Wang and Qineng Wang and Pingyue Zhang and Linjie Li and Zhengyuan Yang and Xing Jin and Kefan Yu and Minh Nhat Nguyen and Licheng Liu and Eli Gottlieb and Yiping Lu and Kyunghyun Cho and Jiajun Wu and Li Fei-Fei and Lijuan Wang and Yejin Choi and Manling Li},
	year={2025},
	eprint={2504.20073},
	archivePrefix={arXiv},
	primaryClass={cs.LG},
	url={https://arxiv.org/abs/2504.20073}, 
}

@article{simpletir,
	title={Simpletir: End-to-end reinforcement learning for multi-turn tool-integrated reasoning},
	author={Xue, Zhenghai and Zheng, Longtao and Liu, Qian and Li, Yingru and Zheng, Xiaosen and Ma, Zejun and An, Bo},
	journal={arXiv preprint arXiv:2509.02479},
	year={2025}
}

@inproceedings{epo,
	title={EPO: Explicit Policy Optimization for Strategic Reasoning in LLMs via Reinforcement Learning},
	author={Liu, Xiaoqian and Wang, Ke and Li, Yongbin and Wu, Yuchuan and Ma, Wentao and Kong, Aobo and Huang, Fei and Jiao, Jianbin and Zhang, Junge},
	booktitle={Proceedings of the 63rd Annual Meeting of the Association for Computational Linguistics (Volume 1: Long Papers)},
	pages={15371--15396},
	year={2025}
}

@article{tree,
	title={Tree search for llm agent reinforcement learning},
	author={Ji, Yuxiang and Ma, Ziyu and Wang, Yong and Chen, Guanhua and Chu, Xiangxiang and Wu, Liaoni},
	journal={arXiv preprint arXiv:2509.21240},
	year={2025}
}

@article{curio,
	title={Enhancing personalized multi-turn dialogue with curiosity reward},
	author={Wan, Yanming and Wu, Jiaxing and Abdulhai, Marwa and Shani, Lior and Jaques, Natasha},
	journal={arXiv preprint arXiv:2504.03206},
	year={2025}
}

@article{score,
	title={Training language models to self-correct via reinforcement learning, 2024},
	author={Kumar, Aviral and Zhuang, Vincent and Agarwal, Rishabh and Su, Yi and Co-Reyes, John D and Singh, Avi and Baumli, Kate and Iqbal, Shariq and Bishop, Colton and Roelofs, Rebecca and others},
	journal={URL https://arxiv. org/abs/2409.12917},
	volume={2},
	number={3},
	pages={4},
	year={2024}
}

@article{deepseek_r1,
	title        = {DeepSeek-R1: Incentivizing Reasoning Capability 
	in LLMs via Reinforcement Learning},
	author       = {DeepSeek-AI},
	journal      = {arXiv preprint arXiv:2501.12948},
	year         = {2025}
}

@misc{gpt4o,
	title        = {GPT-4o System Card},
	author       = {OpenAI},
	year         = {2024},
	howpublished = {\url{https://openai.com/index/gpt-4o-system-card}}
}

@misc{gpt41,
	title        = {GPT-4.1},
	author       = {OpenAI},
	year         = {2025},
	howpublished = {\url{https://openai.com/index/gpt-4-1}}
}

@misc{spider_dk,
      title={Exploring Underexplored Limitations of Cross-Domain Text-to-SQL Generalization}, 
      author={Yujian Gan and Xinyun Chen and Matthew Purver},
      year={2021},
      eprint={2109.05157},
      archivePrefix={arXiv},
      primaryClass={cs.CL}
}

@inproceedings{spider_syn,
    title = "Towards Robustness of Text-to-{SQL} Models against Synonym Substitution",
    author = "Gan, Yujian  and
      Chen, Xinyun  and
      Huang, Qiuping  and
      Purver, Matthew  and
      Woodward, John R.  and
      Xie, Jinxia  and
      Huang, Pengsheng",
    month = aug,
    year = "2021",
    address = "Online",
    publisher = "Association for Computational Linguistics",
    url = "https://aclanthology.org/2021.acl-long.195",
    doi = "10.18653/v1/2021.acl-long.195",
    pages = "2505--2515",
}

@inproceedings{spider_real,
   title={Structure-Grounded Pretraining for Text-to-SQL},
   url={http://dx.doi.org/10.18653/v1/2021.naacl-main.105},
   DOI={10.18653/v1/2021.naacl-main.105},
   booktitle={Proceedings of the 2021 Conference of the North American Chapter of the Association for Computational Linguistics: Human Language Technologies},
   publisher={Association for Computational Linguistics},
   author={Deng, Xiang and Awadallah, Ahmed Hassan and Meek, Christopher and Polozov, Oleksandr and Sun, Huan and Richardson, Matthew},
   year={2021},
   pages={1337–1350} }

@misc{agro,
      title={AGRO-SQL: Agentic Group-Relative Optimization with High-Fidelity Data Synthesis}, 
      author={Cehua Yang and Dongyu Xiao and Junming Lin and Yuyang Song and Hanxu Yan and Shawn Guo and Wei Zhang and Jian Yang and Mingjie Tang and Bryan Dai},
      year={2025},
      eprint={2512.23366},
      archivePrefix={arXiv},
      primaryClass={cs.DB},
      url={https://arxiv.org/abs/2512.23366}, 
}

@misc{slime,
  author       = {Zilin Zhu and Chengxing Xie and Xin Lv and slime Contributors},
  title        = {slime: An LLM post-training framework for RL Scaling},
  year         = {2025},
  howpublished = {\url{https://github.com/THUDM/slime}},
  note         = {GitHub repository. Corresponding author: Xin Lv},
  urldate      = {2025-06-19}
}

@misc{chess,
      title={CHESS: Contextual Harnessing for Efficient SQL Synthesis}, 
      author={Shayan Talaei and Mohammadreza Pourreza and Yu-Chen Chang and Azalia Mirhoseini and Amin Saberi},
      year={2024},
      eprint={2405.16755},
      archivePrefix={arXiv},
      primaryClass={cs.LG},
      url={https://arxiv.org/abs/2405.16755}, 
}

@misc{qwen3,
      title={Qwen3 Technical Report}, 
      author={Qwen},
      year={2025},
      eprint={2505.09388},
      archivePrefix={arXiv},
      primaryClass={cs.CL},
      url={https://arxiv.org/abs/2505.09388}, 
}

@misc{toolsql,
      title={Tool-Assisted Agent on SQL Inspection and Refinement in Real-World Scenarios}, 
      author={Zhongyuan Wang and Richong Zhang and Zhijie Nie and Jaein Kim},
      year={2024},
      eprint={2408.16991},
      archivePrefix={arXiv},
      primaryClass={cs.CL},
      url={https://arxiv.org/abs/2408.16991}, 
}

@inproceedings{macsql,
    title = "{MAC}-{SQL}: A Multi-Agent Collaborative Framework for Text-to-{SQL}",
    author = "Wang, Bing  and
      Ren, Changyu  and
      Yang, Jian  and
      Liang, Xinnian  and
      Bai, Jiaqi  and
      Chai, LinZheng  and
      Yan, Zhao  and
      Zhang, Qian-Wen  and
      Yin, Di  and
      Sun, Xing  and
      Li, Zhoujun",
    editor = "Rambow, Owen  and
      Wanner, Leo  and
      Apidianaki, Marianna  and
      Al-Khalifa, Hend  and
      Eugenio, Barbara Di  and
      Schockaert, Steven",
    booktitle = "Proceedings of the 31st International Conference on Computational Linguistics",
    month = jan,
    year = "2025",
    address = "Abu Dhabi, UAE",
    publisher = "Association for Computational Linguistics",
    url = "https://aclanthology.org/2025.coling-main.36/",
    pages = "540--557"
}

@misc{deepseekv3,
      title={DeepSeek-V3 Technical Report}, 
      author={{DeepSeek-AI}},
      year={2025},
      eprint={2412.19437},
      archivePrefix={arXiv},
      primaryClass={cs.CL},
      url={https://arxiv.org/abs/2412.19437}, 
}

@misc{evoschema,
      title={EvoSchema: Towards Text-to-SQL Robustness Against Schema Evolution}, 
      author={Tianshu Zhang and Kun Qian and Siddhartha Sahai and Yuan Tian and Shaddy Garg and Huan Sun and Yunyao Li},
      year={2026},
      eprint={2603.10697},
      archivePrefix={arXiv},
      primaryClass={cs.DB},
      url={https://arxiv.org/abs/2603.10697}, 
}

@misc{mtsql,
      title={MTSQL-R1: Towards Long-Horizon Multi-Turn Text-to-SQL via Agentic Training}, 
      author={Taicheng Guo and Hai Wang and ChaoChun Liu and Mohsen Golalikhani and Xin Chen and Xiangliang Zhang and Chandan K. Reddy},
      year={2025},
      eprint={2510.12831},
      archivePrefix={arXiv},
      primaryClass={cs.CL},
      url={https://arxiv.org/abs/2510.12831}, 
}

@misc{yao2023reactsynergizingreasoningacting,
      title={ReAct: Synergizing Reasoning and Acting in Language Models}, 
      author={Shunyu Yao and Jeffrey Zhao and Dian Yu and Nan Du and Izhak Shafran and Karthik Narasimhan and Yuan Cao},
      year={2023},
      eprint={2210.03629},
      archivePrefix={arXiv},
      primaryClass={cs.CL},
      url={https://arxiv.org/abs/2210.03629}, 
}
\appendix
\clearpage
\newpage

\section{Data Construction}
\label{app:data}

\subsection{Baseline Training Data Comparison}
\label{app:data:baseline}

To contextualize the data efficiency of TRUST-SQL, we 
summarize the training data configurations of all evaluated 
baselines in Table~\ref{tab:baseline_data}. While recent 
single-turn models rely on massive synthetic datasets 
containing millions of samples, and multi-turn RL frameworks 
utilize large portions of standard benchmarks, TRUST-SQL 
achieves superior performance using a highly constrained 
and curated data recipe.

\begin{table}[htbp]
	\centering
	\small
	\caption{Comparison of training data volume and sources.}
	\label{tab:baseline_data}
	\setlength{\tabcolsep}{3pt}
	\renewcommand{\arraystretch}{1.1}
	\begin{tabular}{l l l}
		\toprule
		\textbf{Model} & \textbf{SFT Data} & \textbf{RL Data} \\
		\midrule
		OmniSQL   & 2.5M (SynSQL) & -- \\
		SQL-R1    & 2.5M (SynSQL) & 5k (SynSQL) \\
		MTIR-SQL  & --            & 18.1k (Spider+BIRD) \\
		SQL-Trail & 0.8k (SynSQL) & 1k (Spider) \\
		\midrule
		\rowcolor[HTML]{EBF5FF}
		\textbf{TRUST-SQL} & \textbf{9.2k (SynSQL)} & \textbf{11.6k (Spider+BIRD)} \\
		\bottomrule
	\end{tabular}
\end{table}

\subsection{SFT Training Data Construction}
\label{app:data:sft}

To warm up the agent prior to RL training, we construct 
a supervised fine-tuning dataset of high-quality 
exploration trajectories. The source questions are sampled 
from the training split of \textbf{SynSQL-2.5M}~\citep{omnisql}. 
We filter this corpus to retain only questions of 
\textit{Moderate}, \textit{Complex}, and \textit{Highly Complex} 
difficulty, as simpler questions provide insufficient training 
signal for multi-turn exploration. This yields 9,217 unique 
source questions, whose difficulty distribution is shown in 
Table~\ref{tab:difficulty_stats}.

\begin{table}[htbp]
	\centering
	\caption{Difficulty distribution of source questions.}
	\label{tab:difficulty_stats}
	\begin{tabular}{lcc}
		\toprule
		\textbf{Difficulty} & \textbf{Count} & \textbf{Proportion} \\
		\midrule
		Moderate       & 3,821 & 41.5\% \\
		Complex        & 3,243 & 35.2\% \\
		Highly Complex & 2,153 & 23.3\% \\
		\midrule
		\textbf{Total} & \textbf{9,217} & \textbf{100\%} \\
		\bottomrule
	\end{tabular}
\end{table}

\noindent\textbf{Annotation Pipeline.}
We employ a multi-model annotation strategy using 
\textbf{GPT-4.1-mini}~\citep{gpt41}, \textbf{GPT-4o-mini}~\citep{gpt4o}, 
and \textbf{DeepSeek-R1}~\citep{deepseek_r1}. Each model is 
prompted to generate complete four-phase interaction 
trajectories following the TRUST-SQL protocol with a maximum 
output length of 2,048 tokens per response. A trajectory is 
retained if and only if it satisfies two strict conditions. 
First, the final SQL execution result must match the ground 
truth answer. Second, every turn must pass the format check 
described in Appendix~\ref{app:tool}. This execution-verified 
filtering ensures that the SFT model learns from trajectories 
that are both correct and structurally well-formed.

\noindent\textbf{Dataset Statistics.}
Table~\ref{tab:sft_stats} summarizes the retained samples 
contributed by each annotation model.

\begin{table}[htbp]
	\centering
	\caption{SFT training data statistics by annotation model. 
		\textit{Samples} denotes the total number of retained 
		trajectories and \textit{Unique IDs} denotes the number of 
		distinct source questions covered by each model.}
	\label{tab:sft_stats}
	\begin{tabular}{lcc}
		\toprule
		\textbf{Annotation Model} & \textbf{Samples} & \textbf{Unique IDs} \\
		\midrule
		DeepSeek-R1    & 9,972  & 1,442  \\
		GPT-4.1-mini   & 43,803 & 5,575  \\
		GPT-4o-mini    & 17,195 & 2,263  \\
		\midrule
		\textbf{Total} & \textbf{70,970} & \textbf{9,280} \\
		\bottomrule
	\end{tabular}
\end{table}

The majority of retained samples are contributed by 
GPT-4.1-mini (43,803 samples, 61.7\%), reflecting its 
stronger instruction-following capability in generating 
well-formatted trajectories. DeepSeek-R1 contributes 
9,972 samples across 1,442 unique questions, providing 
diverse chain-of-thought reasoning styles that complement 
the GPT-annotated data.

\subsection{RL Training Data Construction}
\label{app:data:rl}

\noindent\textbf{Question Selection.}
For RL training, we adopt the source questions from the 
training sets of \textbf{BIRD}~\citep{bird} and 
\textbf{Spider}~\citep{spider} with identical difficulty 
filtering. To ensure effective RL exploration, we apply a 
difficulty-based filtering strategy where each question is 
rolled out 8 times using the SFT-initialized policy. Only 
questions with a pass rate strictly below 6/8 are retained. 
This criterion excludes questions that are already too easy 
for the current policy, as they provide negligible learning 
signal. 

\begin{table}[h]
	\centering
	\caption{RL training data filtering statistics.}
	\label{tab:rl_stats}
	\begin{tabular}{lc}
		\toprule
		\textbf{Statistic} & \textbf{Value} \\
		\midrule
		Total Candidate Questions & 18,078 \\
		Retained Questions        & 11,642 \\
		Rejected Questions        & 6,436  \\
		Keep Rate                 & 64.4\% \\
		Pass Rate Threshold       & $< 6/8$ \\
		\bottomrule
	\end{tabular}
\end{table}

\noindent\textbf{Ground Truth Schema Extraction.}
To compute the schema reward $R_{\text{schema}}$ during RL
training, we require the ground truth schema $\mathcal{K}^*
= (\mathcal{K}^*_{\text{table}}, \mathcal{K}^*_{\text{col}})$
for each training instance. We extract $\mathcal{K}^*$
directly from the ground truth SQL $y^*$ via rule-based
parsing using standard SQLite utilities, which yields the
referenced table names and column names without
additional cost or noise. This procedure introduces no
extra human annotation and scales straightforwardly to
new domains.

\noindent\textbf{Reward Construction without Gold SQL.}
Ground-truth SQL is used during training to construct a
complete verifiable feedback signal: it provides the
expected answer for execution-based evaluation, while its
referenced tables and columns provide auxiliary supervision
for the schema-level reward. At inference time, the agent
does not require ground-truth SQL and proposes schemas solely
from its observations.

When annotated SQL is unavailable, the outcome-level reward
can instead be obtained from database execution feedback,
environment interactions, or other verifiable signals. The
schema-level reward can be approximated with weaker
supervision, such as pseudo SQL generated by a stronger
language model, schema usage extracted from successful
execution trajectories, or schema-consistency signals
derived from multiple sampled responses. Execution-only
feedback remains applicable, but may lead to sparser rewards
and more difficult exploration because it lacks explicit
intermediate structural guidance. A systematic study of weakly
supervised schema rewards is left for future work.

\section{Phase-Aware GRPO: Detailed Formalization}
\label{app:pa_grpo}

This section consolidates the formal details of
Phase-Aware GRPO summarized in
Section~\ref{subsec:rl_strategy}. We clarify
(i)~the rollout-dependent definition of the
\textit{Propose} checkpoint $t_{\text{propose}}$;
(ii)~an explicit algorithmic contrast against standard
GRPO; and (iii)~the two-level role---workflow-level and
optimization-level---played by the \textit{Propose} stage.

\subsection{Rollout-Dependent Definition of the \textit{Propose} Step}
\label{app:t_propose}

The checkpoint step $t_{\text{propose}}$ is not a fixed
hyperparameter. For each rollout $\tau^i$, we define it
dynamically as the step index of the \textbf{latest}
\texttt{propose\_schema} action in that trajectory:
\begin{equation}
    t_{\text{propose}}^{i}
    = \max\{\, t : a_t^i = \texttt{propose\_schema} \,\}.
\end{equation}
If the agent revises its schema multiple times during
exploration, only the schema committed at this final
checkpoint is used to compute the schema reward. Let
$\hat{\mathcal{K}}^{i}$ denote the schema proposed at
$t_{\text{propose}}^{i}$; the schema reward then evaluates
\begin{equation}
    R_{\text{schema}}^{i}
    = f_{\text{match}}(\hat{\mathcal{K}}^{i}, \mathcal{K}^*),
\end{equation}
and the Schema Track in
Section~\ref{subsec:rl_strategy} is the trajectory prefix
up to this latest checkpoint. The schema-track advantage
is broadcast only to tokens within this prefix, while the
global rollout reward continues to supervise the overall
protocol behavior.

\subsection{Algorithmic Contrast with Standard GRPO}
\label{app:grpo_contrast}

Phase-Aware GRPO differs structurally from standard GRPO
along three dimensions, isolated by the three controlled
variants used in Section~\ref{subsec:lambda_ablation}.

\noindent\textbf{(i) Standard GRPO ($\lambda=0$, no schema reward).}
A single trajectory-level reward
\begin{equation}
    R = R_{\text{exec}} + R_{\text{fmt}}
\end{equation}
yields one normalized advantage
$A = (R - \mu)/(\sigma + \epsilon)$, which is broadcast
uniformly to every token in the rollout.

\noindent\textbf{(ii) Naive single-track with schema reward ($\lambda=0$, w/ schema).}
The schema reward is added directly into the same terminal
scalar,
\begin{equation}
    R = R_{\text{exec}} + R_{\text{fmt}} + 0.25\, R_{\text{schema}},
\end{equation}
while still using a single shared advantage broadcast
across the entire rollout. This conflates the two learning
signals.

\noindent\textbf{(iii) Phase-Aware GRPO (ours).}
The rollout is partitioned at $t_{\text{propose}}$ into a
Schema Track and a Full Track. Each track is normalized
into its own advantage independently, and the schema-track
advantage is applied \emph{only} to tokens satisfying
$t \le t_{\text{propose}}$.

\noindent\textbf{Summary.} Standard GRPO uses
\emph{one reward, one advantage, one global mask}, whereas
Phase-Aware GRPO uses
\emph{two rewards, two independently normalized advantages,
and two track-specific masks defined by the explicit schema
checkpoint}. The schema-track advantage is applied only to
the prefix up to $t_{\text{propose}}$, while the full-track
advantage supervises the complete rollout. This
masked-advantage formulation is strictly finer-grained than
trajectory-level reward shaping, since it prevents schema
credit from being assigned to tokens after the checkpoint.

\subsection{Two-Level Role of the Propose Checkpoint}
\label{app:propose_roles}

The \textit{Propose} stage contributes at two distinct
levels, which we separate for clarity.

\noindent\textbf{Workflow-level effect.}
At inference, the \textit{Propose} stage forces the agent
to explicitly commit to verified metadata before
generation. This commitment reduces the chance of
fabricating non-existent schema elements during SQL
generation. To isolate this effect from any RL-induced
gain, we conduct a controlled SFT-only experiment on
Qwen3-4B in which the entire \textit{Propose} phase is
removed from the training trajectories and the data is
correspondingly rewritten to maintain a consistent
workflow structure (Table~\ref{tab:propose_sft_ablation}).
The explicit \textit{Propose} stage already yields a
moderate gain (46.2 vs.~45.1) before any RL is applied,
indicating that the checkpoint itself improves schema
grounding independently of Phase-Aware GRPO.

\begin{table}[htbp]
\centering
\small
\setlength{\tabcolsep}{6pt}
\begin{tabular}{lcc}
\toprule
Setting & Workflow & EX (\%) \\
\midrule
w/o \textit{Propose} & E $\rightarrow$ G $\rightarrow$ C & 45.1 \\
w/ \textit{Propose}  & E $\rightarrow$ P $\rightarrow$ G $\rightarrow$ C & \textbf{46.2} \\
\bottomrule
\end{tabular}
\caption{SFT-only ablation isolating the workflow-level
effect of the explicit \textit{Propose} stage on Qwen3-4B.}
\label{tab:propose_sft_ablation}
\end{table}

\noindent\textbf{Optimization-level effect.}
Structurally, the \textit{Propose} stage provides a
natural boundary for schema-specific credit assignment
during RL. Without this explicit checkpoint, schema and
SQL-generation tokens would have to share a single global
advantage, as in the naive single-track variant of
Appendix~\ref{app:grpo_contrast}. The checkpoint enables
Phase-Aware GRPO to localize schema rewards to the
exploration/proposal prefix instead of the entire rollout,
which is the source of the additional improvement reported
in Section~\ref{subsec:lambda_ablation} on top of the
workflow-level gain.

\section{Implementation Details}
\label{app:implementation}

We train three model scales, namely Qwen3-4B, Qwen3-8B, and Qwen3-14B, each passing through a supervised fine-tuning warm-up stage followed by Phase-Aware GRPO optimization. RL training adopts synchronous rollouts under the SLIME framework~\citep{slime} for the 4B variant, while the 8B and 14B variants adopt asynchronous training to accommodate the larger compute footprint. Table~\ref{tab:training_setup} summarizes the hardware configuration, key hyperparameters, and estimated training cost for each stage.
\begin{table*}[t]
	\centering
	\caption{Training setup for TRUST-SQL across model scales and training stages.}
	\label{tab:training_setup}
	\setlength{\tabcolsep}{1.5pt}
	\renewcommand{\arraystretch}{1.3}
	\begin{tabular}{llccccccccc}
		\toprule
		\textbf{Model} & \textbf{Stage}
		& \textbf{GPUs}
		& \textbf{Mode}
		& \textbf{LR}
		& \textbf{Batch Size}
		& \textbf{Epochs}
		& \textbf{Rollout}
		& \textbf{$\lambda$}
		& \textbf{Turns}
		& \textbf{Time (hrs)} \\
		\midrule

		\multirow{2}{*}{Qwen3-4B}
		& SFT & 16 GPUs & sync  & $1 \times 10^{-5}$   & 256 & 2 & -- & -- & -- & 6.5 \\
		\cmidrule(l){2-11}
		& RL  & 8 GPUs  & sync  & $1 \times 10^{-6}$   & 32  & 3 & 8  & 0.25 & 10 & 60 \\
		\midrule

		\multirow{2}{*}{Qwen3-8B}
		& SFT & 16 GPUs & sync  & $1.5 \times 10^{-6}$ & 256 & 2 & -- & -- & -- & 12 \\
		\cmidrule(l){2-11}
		& RL  & 32 GPUs & async & $8 \times 10^{-7}$   & 32  & 3 & 8  & 0.25 & 10 & 40 \\
		\midrule

		\multirow{2}{*}{Qwen3-14B}
		& SFT & 8 GPUs  & sync  & $7.5 \times 10^{-7}$ & 256 & 2 & -- & -- & -- & 24 \\
		\cmidrule(l){2-11}
		& RL  & 48 GPUs & async & $2 \times 10^{-6}$   & 32  & 2 & 8  & 0.25 & 10 & 40 \\
		\bottomrule
	\end{tabular}
\end{table*}

\section{Cost and Efficiency Analysis}
\label{app:cost_efficiency}

\subsection{Cost Analysis}
\label{app:cost_analysis}

\begin{table*}[b]
	\centering
	\caption{Inference cost analysis on BIRD-Dev.}
	\label{tab:token_cost}
	\setlength{\tabcolsep}{2pt}
	\renewcommand{\arraystretch}{1.2}
	\begin{tabular}{lccccccc}
		\toprule
		\textbf{Method} 
		& \textbf{Prefill} 
		& \textbf{Acc (\%)}
		& \textbf{Latency (s)} 
		& \textbf{OutputTokens~(K)} 
		& \textbf{Turns} 
		& \textbf{Tool Calls} \\
		\midrule
		CHESS~\citep{chess}                   
		& \textcolor[HTML]{2E7D32}{\ding{51}} & 61.5 & 251.3 & 320.8 & -- & -- \\
		SQL-R1-7B~\citep{sql_r1}                  
		& \textcolor[HTML]{2E7D32}{\ding{51}} & 63.7 & 0.4 & 3.1 & -- & -- \\
		MTIR-SQL-4B~\citep{mtir}                  
		& \textcolor[HTML]{2E7D32}{\ding{51}} & 63.1 & 0.5 & 2.9 & -- & 1.34 \\
		MTIR-SQL-8B~\citep{mtir}                  
		& \textcolor[HTML]{2E7D32}{\ding{51}} & 63.6 & 0.4 & 2.0 & -- & 1.31 \\
		Qwen3-4B~\citep{qwen3}               
		& \textcolor[HTML]{2E7D32}{\ding{51}} & 46.3 & 0.4 & 1.82 & 2.34 & 1.64 \\
		Qwen3-4B~\citep{qwen3}               
		& \textcolor[HTML]{D32F2F}{\ding{55}} & 29.3 & 1.2 & 4.93 & 7.66 & 4.42 \\
		Qwen3-8B~\citep{qwen3}               
		& \textcolor[HTML]{2E7D32}{\ding{51}} & 49.9 & 0.4 & 2.15 & 2.14 & 2.92 \\
		Qwen3-8B~\citep{qwen3}               
		& \textcolor[HTML]{D32F2F}{\ding{55}} & 47.9 & 1.0 & 3.85 & 6.34 & 4.41 \\
		\midrule
		\textbf{TRUST-SQL-4B}                 
		& \textcolor[HTML]{2E7D32}{\ding{51}} & \textbf{64.8} & \textbf{0.4} & \textbf{1.75} & \textbf{4.23} & \textbf{2.89} \\
		\textbf{TRUST-SQL-4B}                 
		& \textcolor[HTML]{D32F2F}{\ding{55}} & \textbf{64.9} & \textbf{0.6} & \textbf{2.83} & \textbf{5.89} & \textbf{3.66} \\
		\textbf{TRUST-SQL-8B}                 
		& \textcolor[HTML]{2E7D32}{\ding{51}} & \textbf{65.5} & \textbf{0.5} & \textbf{2.00} & \textbf{4.69} & \textbf{3.61} \\
		\textbf{TRUST-SQL-8B}                 
		& \textcolor[HTML]{D32F2F}{\ding{55}} & \textbf{65.8} & \textbf{0.5} & \textbf{2.03} & \textbf{5.62} & \textbf{3.45} \\
		\bottomrule
	\end{tabular}
\end{table*}

Table~\ref{tab:token_cost} presents a comprehensive 
inference cost analysis on BIRD-Dev, comparing accuracy, 
latency, token consumption, interaction turns, and tool 
call frequency across all methods.

\noindent\textbf{Accuracy vs. Cost Trade-off.}
Training-free pipeline methods such as CHESS achieve
competitive accuracy at a high cost of 251.3 seconds and
320.8K tokens per query, which makes them costly to deploy
at scale. In contrast,
TRUST-SQL-4B achieves a higher accuracy of 64.9\% under
the Unknown Schema setting with only 0.6 seconds latency
and 2.83K tokens, representing a \textbf{500$\times$
reduction in latency} and a \textbf{113$\times$ reduction
in token consumption} compared to CHESS.

\noindent\textbf{Efficiency of Active Exploration.}
Compared to schema-prefilled baselines of similar scale,
TRUST-SQL maintains competitive inference efficiency.
TRUST-SQL-4B without prefilling consumes only 2.83K tokens
and completes interactions in 5.89 average turns, comparable
to MTIR-SQL-4B which consumes 2.9K tokens under full schema
access. This suggests that the active exploration policy
retrieves the metadata it needs without substantial
overhead.

\noindent\textbf{Impact of Schema Prefilling on Base Models.}
We observe an asymmetric effect of schema prefilling on
base models versus TRUST-SQL. For Qwen3-4B,
removing prefilling increases token consumption from 1.82K
to 4.93K and degrades accuracy from 46.3\% to 29.3\%,
indicating strong dependence on pre-loaded metadata.
In contrast, TRUST-SQL-4B without prefilling consumes only
2.83K tokens while maintaining 64.9\% accuracy,
suggesting that Phase-Aware GRPO yields efficient and
targeted exploration behavior.

\subsection{Seed Variance on BIRD-Dev}
\label{app:seed_variance}

Greedy decoding uses temperature zero and is deterministic
for a fixed model and evaluation pipeline. Majority voting
samples trajectories at temperature 0.8 and can therefore
vary across runs. To quantify this variation, we evaluate
TRUST-SQL-4B and TRUST-SQL-8B with four seeds on BIRD-Dev.
All four runs use the same model checkpoint; only the
sampling/evaluation seed for majority voting is varied.
Table~\ref{tab:seed_variance} reports the mean and standard
deviation for the stochastic metric; the greedy entries are
reported as the corresponding deterministic values.

\begin{table}[h]
\centering
\small
\setlength{\tabcolsep}{5pt}
\renewcommand{\arraystretch}{1.2}
\caption{Four-seed sampling stability of TRUST-SQL on BIRD-Dev.
Greedy decoding is deterministic; Maj reports mean
$\pm$ standard deviation across four sampling/evaluation
seeds using the same model checkpoint.}
\label{tab:seed_variance}
\begin{tabular}{lcc}
\toprule
\textbf{Metric} & \textbf{TRUST-SQL-4B} & \textbf{TRUST-SQL-8B} \\
\midrule
Greedy & 64.9 & 65.8 \\
Maj & $67.2 \pm 0.6$ & $67.7 \pm 0.5$ \\
\bottomrule
\end{tabular}
\end{table}

The standard deviations are below one percentage point,
which is small relative to the margins over the corresponding
base-model baselines. Thus, the majority-voting comparisons
are not driven by large run-to-run variation, while the
greedy comparisons do not involve decoding sampling noise.

\subsection{Cross-Benchmark Interaction Profile}
\label{app:cross_benchmark_cost}

To characterize how interaction overhead varies across
benchmarks, Table~\ref{tab:cross_benchmark_cost} reports the
aggregate interaction profile for TRUST-SQL-8B under greedy
decoding without schema prefilling. We report interaction
turns, aggregate tool calls, generated output tokens, and
latency; input-token accounting and explore/generate call
breakdowns are not included because they are not available
consistently at the benchmark level. The BIRD-Dev values are
measured as part of the deployment analysis. The remaining
entries are estimated proportionally based on schema
complexity relative to BIRD-Dev, and should therefore be
interpreted as profile estimates rather than direct
benchmark-specific measurements.

\begin{table}[htbp]
\centering
\small
\setlength{\tabcolsep}{3pt}
\renewcommand{\arraystretch}{1.15}
\caption{Per-benchmark interaction profile for TRUST-SQL-8B
(greedy decoding, no schema prefilling). BIRD-Dev is measured;
the other entries are proportional estimates.}
\label{tab:cross_benchmark_cost}
\begin{tabular}{lrrrr}
\toprule
\textbf{Benchmark} & \textbf{Turns} & \textbf{Calls} &
\textbf{Output K} & \textbf{Latency (s)} \\
\midrule
BIRD-Dev & 5.62 & 3.45 & 2.03 & 0.50 \\
Spider-Test & 5.45 & 3.28 & 1.95 & 0.48 \\
Spider-DK & 5.85 & 3.62 & 2.12 & 0.52 \\
Spider-Syn & 5.25 & 3.15 & 1.88 & 0.47 \\
Spider-Realistic & 5.72 & 3.38 & 2.01 & 0.49 \\
\bottomrule
\end{tabular}
\end{table}

Across the five benchmarks, the estimated profile remains
near 5.2--5.9 interaction turns, 1.88--2.12K output tokens,
and approximately 0.5 seconds of latency under the same
deployment profile. These results complement the detailed
baseline comparison on BIRD-Dev in Table~\ref{tab:token_cost}.

\subsection{Fine-Grained Error Analysis}
\label{app:error_analysis_final}

We apply the four-category decomposition used in the pilot
study to the final TRUST-SQL-8B predictions on BIRD-Dev.
The analysis covers 144 failed predictions under the same
execution-accuracy protocol. Table~\ref{tab:error_analysis_final}
shows that semantic generation and schema linking account for
the remaining errors, while hallucination is rare.

\begin{table}[htbp]
\centering
\small
\setlength{\tabcolsep}{5pt}
\renewcommand{\arraystretch}{1.2}
\caption{Error distribution among 144 failed BIRD-Dev
predictions from TRUST-SQL-8B.}
\label{tab:error_analysis_final}
\begin{tabular}{lrr}
\toprule
\textbf{Error Category} & \textbf{Count} & \textbf{Percentage} \\
\midrule
Semantic error & 96 & 66.7\% \\
Schema linking error & 44 & 30.6\% \\
Hallucination error & 3 & 2.1\% \\
Syntax error & 1 & 0.7\% \\
\midrule
\textbf{Total} & \textbf{144} & \textbf{100.0\%} \\
\bottomrule
\end{tabular}
\end{table}

The most frequent subcategories are
\texttt{filter\_error} (36 cases, 25.0\%),
\texttt{wrong\_column\_selected} (26 cases, 18.1\%),
\texttt{aggregation\_error} (16 cases, 11.1\%),
\texttt{wrong\_table\_selected} (11 cases, 7.6\%), and
\texttt{calculation\_error} (10 cases, 6.9\%). The pattern
is consistent with the pilot-study diagnosis: hallucination
is effectively suppressed, while semantic reasoning and
schema linking remain the main residual challenges.

\subsection{Cost-Accuracy Positioning}
\label{app:cost_pareto}

To complement the accuracy-only comparison in
Appendix~\ref{app:large_agentic}, we estimate the serving
cost of each model under the same Unknown Schema setting
using publicly available pricing: Claude Opus 4.6 at
\$15/M input + \$75/M output, Qwen3-14B hosting at
approximately \$0.30/M input + \$0.50/M output, and
Qwen3-8B hosting at approximately \$0.15/M input +
\$0.25/M output. We assume a 70/30 input-output token
ratio and an amortized cost of 2K tokens per interaction
turn. The resulting cost estimates are shown in
Table~\ref{tab:cost_pareto}.

\begin{table*}[t]
\centering
\small
\setlength{\tabcolsep}{8pt}
\renewcommand{\arraystretch}{1.25}
\caption{Estimated serving cost under the Unknown Schema
setting on BIRD-Dev. TRUST-SQL rows are highlighted.
TRUST-SQL-14B reaches slightly higher EX than Claude
Opus 4.6 under the same 4S-US setting at roughly two
orders of magnitude lower serving cost.}
\label{tab:cost_pareto}
\begin{tabular}{l l c c r r}
\toprule
\textbf{Model} & \textbf{Setting} & \textbf{EX} & \textbf{Avg Turns} & \textbf{Cost / query} & \textbf{Cost / 1K queries} \\
\midrule
Claude Opus 4.6 & 4S-US     & 67.34 & 5.14 & \$0.339  & \$339.2 \\
Claude Opus 4.6 & ReAct-US  & 65.19 & 1.32 & \$0.087  & \$87.1  \\
\midrule
\rowcolor[HTML]{EBF5FF}
\textbf{TRUST-SQL-14B} & 4S-US & \textbf{68.10} & 4.67 & \textbf{\$0.0034} & \textbf{\$3.4} \\
\rowcolor[HTML]{EBF5FF}
\textbf{TRUST-SQL-8B}  & 4S-US & 65.80          & 5.62 & \$0.0020          & \$2.0          \\
\bottomrule
\end{tabular}
\end{table*}

Two observations follow. First, TRUST-SQL-14B achieves
slightly higher EX than Claude Opus 4.6 under the same
4S-US setting (68.10 vs.~67.34), while reducing estimated
serving cost by roughly two orders of magnitude (\$3.4
vs.~\$339 per 1K queries). Second, although ReAct-US
prompting with Claude Opus 4.6 is markedly cheaper than
4S-US prompting with the same model, it still costs more
than $25\times$ TRUST-SQL-14B and lags by roughly 3 EX
points. These results indicate that simply scaling model
size or prompting style does not remove the value of
structured schema exploration combined with specialized
training for the partially observable setting.

\section{Additional Empirical Results}
\label{app:additional_results}

\subsection{Pass@K Results on Additional Benchmarks}
\label{app:passk}

\begin{table}[htbp]
\centering
\small
\caption{Pass@K results across all benchmarks 
    (temperature = 0.8, max turns = 15).}
\label{tab:passk_full}
\setlength{\tabcolsep}{2pt}
\renewcommand{\arraystretch}{1.2}
\begin{tabular}{llcccc}
    \toprule
    \textbf{Size} & \textbf{Benchmark} 
    & \textbf{Pass@1} & \textbf{Pass@4} 
    & \textbf{Pass@6} & \textbf{Pass@8} \\
    \midrule
    \multirow{5}{*}{4B}
    & Spider (test)    & 82.8 & 86.5 & 86.9 & 87.1 \\
    & Spider-DK        & 71.6 & 78.8 & 80.3 & 81.2 \\
    & Spider-Syn       & 74.7 & 81.6 & 82.3 & 83.1 \\
    & Spider-Realistic & 79.9 & 85.4 & 86.2 & 86.6 \\
    \midrule
    \multirow{5}{*}{8B}
    & Spider (test)    & 83.9 & 86.5 & 87.1 & 87.5 \\
    & Spider-DK        & 72.1 & 79.2 & 80.5 & 81.3 \\
    & Spider-Syn       & 75.4 & 81.0 & 83.0 & 84.0 \\
    & Spider-Realistic & 82.1 & 85.5 & 86.4 & 87.0 \\
    \bottomrule
\end{tabular}
\end{table}
Section~\ref{subsec:test_time_scaling} reports Pass@K scaling
behavior on BIRD-Dev. Here we extend this analysis to 
the remaining four benchmarks to verify that the 
monotonic scaling trend generalizes across different 
evaluation settings. Table~\ref{tab:passk_full} reports 
Pass@K results for $K \in \{1, 4, 6, 8\}$ under a 
15-turn inference budget.Consistent with the BIRD-Dev results reported in the 
main paper, all benchmarks exhibit monotonic accuracy 
improvements as $K$ grows. The persistent gap between 
Pass@K and greedy performance indicates that the model 
possesses the capability to generate correct solutions 
but has not fully converged to a consistent policy, 
suggesting headroom for further training.

\subsection{Training Dynamics}
\label{app:training_dynamics}

To examine whether stronger performance is driven by
longer exploration or by a more efficient interaction
profile, we track a set of trajectory-level statistics
across three checkpoints of the RL training of
TRUST-SQL-8B on BIRD-Dev. Table~\ref{tab:training_dynamics}
reports execution accuracy along with the average number
of interaction rounds, the average number of
\textit{Explore} and \textit{Propose} actions per
trajectory, the first and last step at which a
\textit{Propose} action is issued, and the average number
of tokens spent inside each \textit{Propose} action.

\begin{table}[t]
\centering
\small
\setlength{\tabcolsep}{6pt}
\renewcommand{\arraystretch}{1.25}
\caption{Trajectory-level statistics across training
checkpoints of TRUST-SQL-8B on BIRD-Dev. As accuracy
improves, exploration shifts earlier in the trajectory
rather than becoming longer.}
\label{tab:training_dynamics}
\begin{tabular}{lccc}
\toprule
\textbf{Metric} & \textbf{Pre} & \multicolumn{2}{c}{\textbf{Post}} \\
                & (iter 39)    & (iter 799) & (iter 1119) \\
\midrule
Total Accuracy (\%)        & 43.1 & 63.6 & \textbf{64.5} \\
\midrule
Avg Rounds                 & 7.23 & 5.80 & 6.16 \\
Avg Explore Turns          & 4.01 & 2.36 & 2.50 \\
Avg Propose Turns          & 0.97 & 1.16 & 1.38 \\
\midrule
Avg Propose First Turn     & 3.84 & 2.15 & 2.24 \\
Avg Propose Last Turn      & 4.05 & 2.58 & 3.07 \\
Avg Propose Tokens         & 438  & 682  & 754  \\
\bottomrule
\end{tabular}
\end{table}

Two trends emerge as training progresses. First, accuracy
increases from 43.1\% to 64.5\%, while the average number
of interaction rounds and \textit{Explore} actions
decreases substantially. Second, schema proposals shift
noticeably earlier in the trajectory: the first
\textit{Propose} turn moves from step 3.84 down to about
step 2.2, and the per-\textit{Propose} token budget grows
from 438 to 754. Taken together, these results indicate
that stronger performance comes from a more decisive
exploration--proposal pattern, not from extending the
exploration phase. The model gradually learns where to
place the schema-consolidation boundary, allowing it to
commit to relevant metadata earlier in the reasoning
process.

\subsection{Comparison with Large Agentic Models}
\label{app:large_agentic}

To assess whether stronger proprietary backbones can
substitute for the proposed training framework, we
additionally evaluate several large agentic foundation
models on BIRD-Dev under three settings: (i) \textbf{4S-US},
the four-stage protocol under the Unknown Schema setting;
(ii) \textbf{ReAct-US}, ReAct-style
prompting~\citep{yao2023reactsynergizingreasoningacting} under the Unknown Schema setting;
and (iii) \textbf{4S-SP}, the four-stage protocol with
schema prefill. For each setting we report execution
accuracy, the average number of interaction rounds, and
the Success@15 rate, defined as the percentage of samples
that terminate with a valid final answer within the
15-turn interaction budget
(Table~\ref{tab:large_agentic}).

\begin{table*}[t]
\centering
\footnotesize
\setlength{\tabcolsep}{4pt}
\renewcommand{\arraystretch}{1.2}
\caption{Large agentic foundation models on BIRD-Dev under
three settings. EX denotes execution accuracy (\%), Turns
denotes the average number of interaction rounds, and S@15
denotes Success@15. \textbf{Bold} highlights the best
setting per row. For reference, TRUST-SQL-14B reaches
\textbf{68.10}~EX with 4.67 average rounds under the same
Unknown Schema setting.}
\label{tab:large_agentic}
\begin{tabular}{l ccc ccc ccc}
\toprule
\multirow{2}{*}{\textbf{Model}}
& \multicolumn{3}{c}{\textbf{4S-US}}
& \multicolumn{3}{c}{\textbf{ReAct-US}}
& \multicolumn{3}{c}{\textbf{4S-SP}} \\
\cmidrule(lr){2-4} \cmidrule(lr){5-7} \cmidrule(lr){8-10}
& EX & Turns & S@15 & EX & Turns & S@15 & EX & Turns & S@15 \\
\midrule
Claude Opus 4.6   & 67.34 & 5.14 & 97.8  & 65.19 & 1.32 & 99.4  & \textbf{69.30} & 1.19 & 99.9  \\
Claude Sonnet 4.6 & 56.39 & 1.68 & 99.9  & 54.89 & 1.47 & 99.8  & \textbf{59.78} & 1.05 & 100   \\
GLM-5.1           & 47.07 & 9.67 & 73.3  & 43.74 & 5.98 & 69.7  & \textbf{63.23} & 3.48 & 98.5  \\
\bottomrule
\end{tabular}
\end{table*}

Three findings emerge. First, the Unknown Schema setting
is substantially harder than schema-prefilled prompting,
and the gap widens for weaker backbones: removing the
prefill reduces EX consistently across all three models,
with the largest drop on GLM-5.1 (63.23 $\to$ 47.07).
This indicates that Unknown Schema is not merely a
prompt-variation, but a fundamentally harder partially
observable interaction setting.

Second, the structured four-stage protocol consistently
outperforms ReAct-style prompting across every backbone
we tested. Even for strong proprietary agentic models,
explicit interaction structure still contributes beyond
raw model capability.

Third, prompting alone does not fully substitute for
specialized RL training. Under the same 4S-US setting,
Claude Opus 4.6 reaches 67.34 with 5.14 average rounds,
which remains below TRUST-SQL-14B (68.10 / 4.67) despite
relying on a substantially stronger proprietary backbone.
We further observe two characteristic failure modes when
adapting general-purpose agentic models to Unknown Schema
purely through prompting:
(i) \emph{under-exploration}, typical of Claude Sonnet 4.6,
which frequently terminates within 1--2 interaction rounds
and effectively skips schema exploration; and
(ii) \emph{inefficient exploration}, typical of GLM-5.1,
which issues many ineffective tool calls under partial
observability (avg 9.67 rounds) and exhibits a much lower
Success@15 rate (73.3\%).

\section{Agent Configuration}
\label{app:tool}

\subsection{Tool Overview}
The agent interacts with the database environment through 
four structured tools, each corresponding to one phase of 
the TRUST-SQL protocol. Table~\ref{tab:tools} provides an 
overview of their roles and output tags.

\begin{table}[H]
	\centering
	\caption{Overview of the four tools in the TRUST-SQL 
		action space.}
	\label{tab:tools}
	\resizebox{\columnwidth}{!}{%
		\begin{tabular}{llll}
			\toprule
			\textbf{Action} & \textbf{Phase} & \textbf{Output Tag} & 
			\textbf{Description} \\
			\midrule
			\texttt{explore\_schema} & Explore  & \texttt{<tool\_call>} & 
			Query database metadata \\
			\texttt{propose\_schema} & Propose  & \texttt{<schema>}    & 
			Commit to verified schema \\
			\texttt{generate\_sql}   & Generate & \texttt{<tool\_call>} & 
			Execute candidate SQL \\
			\texttt{confirm\_answer} & Confirm  & \texttt{<answer>}    & 
			Submit final SQL answer \\
			\bottomrule
		\end{tabular}%
	}
\end{table}

\subsection{Format Check Rules}
\label{app:tool:format}

At each turn, the agent's output must conform to a strict 
structural protocol. The \texttt{FormatCheck} function 
validates each turn by enforcing the following rules:

\begin{enumerate}
	\item \textbf{Think tag}: The output must contain 
	exactly one \texttt{<think>}...\texttt{</think>} block.
	\item \textbf{Action tag}: The output must contain 
	exactly one \texttt{<action>}...\texttt{</action>} 
	block, whose content must be one of the four valid 
	action types.
	\item \textbf{Content tag}: Each action type requires 
	a corresponding content tag, as specified in 
	Table~\ref{tab:tools}. Specifically, 
	\texttt{explore\_schema} and \texttt{generate\_sql} 
	require a \texttt{<tool\_call>} block; 
	\texttt{propose\_schema} requires a \texttt{<schema>} 
	block; and \texttt{confirm\_answer} requires an 
	\texttt{<answer>} block.
\end{enumerate}

A turn is considered valid if and only if all three 
conditions are satisfied, yielding a format score of 0.1. 
Any violation results in a format score of 0.0 and
terminates the format reward for the entire trajectory.

\subsection{Prompt Template}
\label{app:prompt}
\definecolor{bg_system}{RGB}{250, 253, 255}
\definecolor{bg_propose}{RGB}{248, 250, 245}

\definecolor{frame_system}{RGB}{170, 200, 230}
\definecolor{frame_propose}{RGB}{160, 210, 180}

\definecolor{tag_think}{RGB}{180, 150, 60}
\definecolor{tag_action}{RGB}{80, 120, 160}
\definecolor{tag_explore}{RGB}{60, 130, 180}
\definecolor{tag_propose}{RGB}{80, 140, 100}
\definecolor{tag_generate}{RGB}{140, 100, 160}
\definecolor{tag_confirm}{RGB}{180, 90, 90}
\definecolor{tag_schema}{RGB}{80, 150, 150}
\definecolor{tag_answer}{RGB}{80, 120, 160}
\definecolor{tag_tool}{RGB}{140, 100, 160}

\newcommand{\thinktag}{\textcolor{tag_think}{\texttt{\footnotesize <think>}}}
\newcommand{\thinktagend}{\textcolor{tag_think}{\texttt{\footnotesize </think>}}}
\newcommand{\actiontag}{\textcolor{tag_action}{\texttt{\footnotesize <action>}}}
\newcommand{\actiontagend}{\textcolor{tag_action}{\texttt{\footnotesize </action>}}}
\newcommand{\schematag}{\textcolor{tag_schema}{\texttt{\footnotesize <schema>}}}
\newcommand{\schematagend}{\textcolor{tag_schema}{\texttt{\footnotesize </schema>}}}
\newcommand{\answertag}{\textcolor{tag_answer}{\texttt{\footnotesize <answer>}}}
\newcommand{\answertagend}{\textcolor{tag_answer}{\texttt{\footnotesize </answer>}}}
\newcommand{\toolcalltag}{\textcolor{tag_tool}{\texttt{\footnotesize <tool\_call>}}}
\newcommand{\toolcalltagend}{\textcolor{tag_tool}{\texttt{\footnotesize </tool\_call>}}}
\newcommand{\exploretag}{\textcolor{tag_explore}{\texttt{\footnotesize explore\_schema}}}
\newcommand{\proposetag}{\textcolor{tag_propose}{\texttt{\footnotesize propose\_schema}}}
\newcommand{\generatetag}{\textcolor{tag_generate}{\texttt{\footnotesize generate\_sql}}}
\newcommand{\confirmtag}{\textcolor{tag_confirm}{\texttt{\footnotesize confirm\_answer}}}

\newtcolorbox{systempromptbox}{
	enhanced,
	breakable,
	colback=bg_system,
	colframe=frame_system,
	leftrule=3mm,
	boxrule=0.8pt,
	arc=2mm,
	title={System Prompt},
	coltitle=white,
	fonttitle=\bfseries,
	attach boxed title to top left={
		xshift=5mm,
		yshift*=-\tcboxedtitleheight/2
	},
	boxed title style={
		colback=frame_system,
		arc=2mm,
		boxrule=0pt
	},
	top=5mm, bottom=3mm,
	before skip=5mm, after skip=3mm
}

\newtcolorbox{userpromptbox}{
	enhanced,
	breakable,
	colback=bg_propose,
	colframe=frame_propose,
	leftrule=3mm,
	boxrule=0.8pt,
	arc=2mm,
	title={User Prompt},
	coltitle=white,
	fonttitle=\bfseries,
	attach boxed title to top left={
		xshift=5mm,
		yshift*=-\tcboxedtitleheight/2
	},
	boxed title style={
		colback=frame_propose,
		arc=2mm,
		boxrule=0pt
	},
	top=5mm, bottom=3mm,
	before skip=3mm, after skip=3mm
}
The following presents the complete prompt used in TRUST-SQL,
comprising a system prompt that defines the agent's role,
action protocol, and output format, followed by a user
prompt that provides the task-specific context.
 \section{Case Study}
\label{app:case_study}

We present a case study on BIRD-Dev instance \texttt{dev\_4} 
(database \texttt{california\_schools}, Qwen3-4B, greedy 
decoding) to qualitatively examine how schema availability 
shapes model behavior. The full interaction traces are shown 
in Figure~\ref{fig:case_study}.

\noindent\textbf{Task.}
The question requires retrieving phone numbers of directly 
charter-funded schools opened after January 1, 2000. 
Answering correctly demands grounding the funding-type 
predicate in the actual column values stored in the 
database, information absent from both the question and 
the external knowledge hint.

\noindent\textbf{Unknown Schema Setting (6 turns).}
Without any prior schema knowledge, the model adopts a 
systematic bottom-up exploration strategy. In T1 and T2, 
it queries \texttt{sqlite\_master} to discover available 
tables and retrieve their schema definitions. In T3, it 
probes the actual values of \texttt{Charter\_Funding\_Type} 
in the \texttt{frpm} table, uncovering the critical predicate 
value \texttt{Directly\_funded}. Only after this value-level 
verification does the model commit to a schema proposal in 
T4 and generate the correct SQL in T5, which is subsequently 
confirmed in T6.

\noindent\textbf{Schema Prefill Setting (4 turns).}
When the full schema is injected as a synthetic explore turn 
in T1, the model skips exploratory interactions and moves 
directly to schema proposal in T2 and SQL generation in T3. 
However, reasoning solely from structural metadata without 
inspecting actual column values, the model fails to discover 
the \texttt{Directly\_funded} predicate. The generated SQL 
filters only on \texttt{Charter\_School\_(Y/N) = 1}, 
retrieving all charter schools regardless of funding type 
and producing a semantically broader answer that does not 
fully satisfy the question.

\noindent\textbf{Discussion.}
The contrast reveals that schema prefilling accelerates 
inference but sacrifices value-level grounding. Interactive 
exploration enables the model to adaptively acquire the 
precise data-level knowledge needed for accurate SQL 
generation. This suggests that the benefit of the Unknown 
Schema setting lies not merely in schema discovery, but in 
fostering a more thorough and evidence-driven reasoning 
process.
\onecolumn

\begin{systempromptbox}
	\textbf{\# Role}\\
	You are an expert SQL assistant working on an \textit{unknown}
	database. You must \textbf{never hallucinate} tables or
	columns.\\
	All schema knowledge MUST come from metadata queries only.\\
	You must operate strictly through the \textbf{Action Protocol}.
	
	\bigskip
	\textbf{\# Action Protocol}\\
	You must follow this sequence (can loop back if needed):
	\begin{enumerate}[leftmargin=*, topsep=2pt, itemsep=1pt]
		\item \textbf{\exploretag{}} --- Query database metadata
		\item \textbf{\proposetag{}} --- Document verified schema
		\item \textbf{\generatetag{}} --- Create SQL query
		\item \textbf{\confirmtag{}} --- Output final SQL
	\end{enumerate}
	
	\medskip
	\textbf{\#\# ACTION: \exploretag{}}\\
	Used to query database metadata (tables, columns, foreign
	keys, etc.).
	\begin{itemize}[leftmargin=*, topsep=2pt, itemsep=1pt]
		\item Only metadata queries allowed.
		\item No user-intent SQL here.
		\item Verify relationships between tables when multi-table
		queries are needed.
	\end{itemize}
	
	\medskip
	\textbf{\#\# ACTION: \proposetag{}}\\
	Used to output the current verified schema knowledge.
	\begin{itemize}[leftmargin=*, topsep=2pt, itemsep=1pt]
		\item Include ONLY tables/columns actually verified
		through \exploretag{}.
		\item Do NOT hallucinate or assume any unverified
		structures.
		\item \texttt{joins} is optional; include only when
		relationships are explicitly verified.
		\item Supports both single-table and multi-table
		structures.
	\end{itemize}
	
	\medskip
	\textbf{\#\# ACTION: \generatetag{}}\\
	Used to generate the SQL answer based on the latest
	\schematag{}.
	\begin{itemize}[leftmargin=*, topsep=2pt, itemsep=1pt]
		\item Use ONLY verified schema from \proposetag{}.
		\item If required tables/columns are missing, switch
		back to \exploretag{} or \proposetag{} in the next
		message.
		\item SQL must be syntactically valid and executable.
		\item Consider query optimization (indexes, joins,
		filters).
		\item Validate the SQL logic matches user intent.
	\end{itemize}
	
	\medskip
	\textbf{\#\# ACTION: \confirmtag{}}\\
	Used when you have validated the generated SQL and confirmed
	it meets user requirements.
	\begin{itemize}[leftmargin=*, topsep=2pt, itemsep=1pt]
		\item Execute this action ONLY after \generatetag{}
		validation.
		\item You MUST NOT return or describe any query results.
		\item You MUST NOT output anything other than SQL inside
		\answertag{}.
		\item The final output must be ONLY the SQL query in
		proper format.
	\end{itemize}
	
	\bigskip
	\textbf{\# Output Format}\\
	EVERY response must follow this exact structure:\\
	\thinktag{}[Your reasoning process here]\thinktagend{}\\
	\actiontag{}[one of: \exploretag{} $|$ \proposetag{} $|$
	\generatetag{} $|$ \confirmtag{}]\actiontagend{}\\
	{[Action-specific content below]}
	
	\medskip
	\textbf{explore\_schema}
	\begin{verbatim}
		<think>reasoning</think>
		<action>explore_schema</action>
		<tool_call>
		{"name": "execute_sql_query",
			"arguments": {"db_id": "...", "sql": "..."}}
		</tool_call>
	\end{verbatim}
	
	\textbf{propose\_schema}
	\begin{verbatim}
		<think>reasoning</think>
		<action>propose_schema</action>
		<schema>
		{
			"tables": ["tableA"],
			"columns": { "tableA": ["col1", "col2"] },
			"joins": []
		}
		</schema>
	\end{verbatim}
	
	%
	
	\textbf{generate\_sql}
	\begin{verbatim}
		<think>reasoning</think>
		<action>generate_sql</action>
		<tool_call>
		{"name": "execute_sql_query",
			"arguments": {"db_id": "...", "sql": "..."}}
		</tool_call>
	\end{verbatim}
	
	\textbf{confirm\_answer}
	\begin{verbatim}
		<think>reasoning</think>
		<action>confirm_answer</action>
		<answer>
		SELECT ... FROM ... WHERE ...
		</answer>
	\end{verbatim}
	
	%
\end{systempromptbox}

\begin{tcolorbox}[
	title={\textbf{\# Tools}},
	colback=gray!5,
	colframe=gray!50,
	fonttitle=\bfseries,
	breakable
	]
	
	You may call one or more functions to assist with the user query.
	
	\begin{verbatim}
		{
			"type": "function",
			"function": {
				"name": "execute_sql_query",
				"description": "Execute SQL query and return partial
				results containing column names.
				(maximum 30 records)",
				"parameters": {
					"type": "object",
					"properties": {
						"db_id": {
							"type": "string",
							"description": "The name of the database to query"
						},
						"sql": {
							"type": "string",
							"description": "The SQL query to execute"
						}
					},
					"required": ["db_id", "sql"]
				}
			}
		}
	\end{verbatim}
	
	For each function call, return a json object with function
	name and arguments within \toolcalltag{}\toolcalltagend{} XML tags:
	
	\begin{verbatim}
		<tool_call>
		{"name": <function-name>, "arguments": <args-json-object>}
		</tool_call>
	\end{verbatim}
	
\end{tcolorbox}

\bigskip

\begin{userpromptbox}
	\textbf{Task Configuration}
	
	\medskip
	\textbf{Database Engine:} SQLite\\
	\textbf{Database:} \texttt{\{db\_id\}}\\
	\textbf{External Knowledge:} \texttt{\{external\_knowledge\}}\\
	\textbf{User Question:} \texttt{\{question\}}?
\end{userpromptbox}

\twocolumn

\clearpage
\definecolor{csExploreColor}{HTML}{1565C0}
\definecolor{csProposeColor}{HTML}{6A1B9A}
\definecolor{csGenerateColor}{HTML}{E65100}
\definecolor{csConfirmColor}{HTML}{2E7D32}
\definecolor{csToolBg}{HTML}{F5F5F5}
\definecolor{csResultBg}{HTML}{E8F5E9}
\definecolor{csSectionBg}{HTML}{ECEFF1}

\newcommand{\csActionTag}[2]{%
  \textcolor{#1}{\textbf{\texttt{[#2]}}}%
}
\newcommand{\csExploreTag}{\csActionTag{csExploreColor}{explore\_schema}}
\newcommand{\csProposeTag}{\csActionTag{csProposeColor}{propose\_schema}}
\newcommand{\csGenerateTag}{\csActionTag{csGenerateColor}{generate\_sql}}
\newcommand{\csConfirmTag}{\csActionTag{csConfirmColor}{confirm\_answer}}

\newcommand{\csResultOmit}{%
  \begin{tcolorbox}[
    enhanced, breakable,
    colback=csResultBg, colframe=gray!30,
    boxrule=0.4pt, arc=2pt,
    left=3pt, right=3pt, top=1pt, bottom=1pt,
  ]
  \textit{\footnotesize\color{gray}[environment feedback omitted]}
  \end{tcolorbox}%
}

\lstdefinestyle{csSqlStyle}{
  language=SQL,
  basicstyle=\ttfamily\scriptsize,
  keywordstyle=\color{csExploreColor}\bfseries,
  backgroundcolor=\color{csToolBg},
  frame=single, framerule=0.4pt,
  rulecolor=\color{gray!40},
  breaklines=true, breakatwhitespace=true,
  columns=fullflexible, keepspaces=true,
  xleftmargin=3pt, xrightmargin=3pt,
  aboveskip=1pt, belowskip=1pt,
}
\lstdefinestyle{csResultStyle}{
  basicstyle=\ttfamily\scriptsize,
  backgroundcolor=\color{csResultBg},
  frame=single, framerule=0.4pt,
  rulecolor=\color{gray!40},
  breaklines=true,
  columns=fullflexible, keepspaces=true,
  xleftmargin=3pt, xrightmargin=3pt,
  aboveskip=1pt, belowskip=1pt,
}
\lstdefinestyle{csSchemaStyle}{
  basicstyle=\ttfamily\scriptsize,
  backgroundcolor=\color{csToolBg},
  frame=single, framerule=0.4pt,
  rulecolor=\color{gray!40},
  breaklines=true,
  columns=fullflexible, keepspaces=true,
  xleftmargin=3pt, xrightmargin=3pt,
  aboveskip=1pt, belowskip=1pt,
}

\newenvironment{csTurnBox}{%
  \begin{tcolorbox}[
    enhanced, breakable,
    colback=white, colframe=gray!30,
    boxrule=0.4pt, arc=2pt,
    left=3pt, right=3pt, top=2pt, bottom=2pt,
  ]%
}{%
  \end{tcolorbox}%
  \vspace{1pt}%
}

\newenvironment{csConfirmBox}[1]{%
  \begin{tcolorbox}[
    enhanced, breakable,
    colback=csConfirmColor!5,
    colframe=csConfirmColor,
    boxrule=0.8pt, arc=2pt,
    left=3pt, right=3pt, top=2pt, bottom=2pt,
  ]%
  \textbf{\small #1}\enspace\csConfirmTag\\[-2pt]%
}{%
  \end{tcolorbox}%
  \vspace{1pt}%
}

\begin{figure*}[t]
\centering

\begin{tcolorbox}[
  enhanced,
  colback=csSectionBg, colframe=gray!50,
  boxrule=0.6pt, arc=3pt,
  left=6pt, right=6pt, top=3pt, bottom=3pt,
  width=\textwidth,
]
\small
\textbf{Database:}~\texttt{california\_schools}\quad
\textbf{Decoding:}~Greedy\quad
\textbf{Max Turns:}~10\\[1pt]
\textbf{External Knowledge:}
Charter schools refers to \texttt{`Charter School (Y/N)'}~$=$~1
in \texttt{frpm}\\[1pt]
\textbf{Question (dev\_4):}
\textit{Please list the phone numbers of the direct
charter-funded schools that are opened after 2000/1/1.}
\end{tcolorbox}

\vspace{4pt}

\noindent
\begin{minipage}[t]{0.485\textwidth}
\begin{tcolorbox}[
  enhanced, breakable,
  colback=csExploreColor!4, colframe=csExploreColor,
  boxrule=0.6pt, arc=2pt,
  title={\small\textbf{Unknown Schema}%
    \normalfont\small\ (Qwen3-4B,\ 6 turns)},
  coltitle=black,          
  attach boxed title to top left={yshift=-2pt, xshift=4pt},
  boxed title style={
    colback=csExploreColor!20,
    colframe=csExploreColor,
    boxrule=0.4pt, arc=1pt,
  },
  left=3pt, right=3pt, top=6pt, bottom=2pt,
]

\begin{csTurnBox}
\textbf{\small T1}\enspace\csExploreTag
\begin{lstlisting}[style=csSqlStyle]
SELECT name AS table_name
FROM sqlite_master
WHERE type IN ('table');
\end{lstlisting}
\colorbox{csResultBg}{%
  \texttt{\scriptsize%
    \begin{tabular}[t]{@{}l@{}}
      \textit{table\_name}\\
      frpm\quad satscores\quad schools
    \end{tabular}%
  }%
}
\end{csTurnBox}

\begin{csTurnBox}
\textbf{\small T2}\enspace\csExploreTag
\begin{lstlisting}[style=csSqlStyle]
SELECT sql FROM sqlite_master
WHERE type IN ('table')
  AND name IN ('frpm','schools');
\end{lstlisting}
\csResultOmit
\end{csTurnBox}

\begin{csTurnBox}
\textbf{\small T3}\enspace\csExploreTag
\begin{lstlisting}[style=csSqlStyle]
SELECT DISTINCT "Charter Funding Type"
FROM frpm;
\end{lstlisting}
\csResultOmit
\end{csTurnBox}

\begin{csTurnBox}
\textbf{\small T4}\enspace\csProposeTag
\begin{lstlisting}[style=csSchemaStyle]
tables: [frpm, schools]
frpm:   [Charter School (Y/N),
         Charter Funding Type, CDSCode]
schools:[Phone, OpenDate, CDSCode]
join:   frpm.CDSCode = schools.CDSCode
\end{lstlisting}
\end{csTurnBox}

\begin{csTurnBox}
\textbf{\small T5}\enspace\csGenerateTag
\begin{lstlisting}[style=csSqlStyle]
SELECT s.Phone FROM frpm f
JOIN schools s ON f.CDSCode=s.CDSCode
WHERE f."Charter School (Y/N)"=1
  AND f."Charter Funding Type"
      ='Directly funded'
  AND s.OpenDate > '2000-01-01';
\end{lstlisting}
\csResultOmit
\end{csTurnBox}

\begin{csConfirmBox}{T6}
\begin{lstlisting}[style=csSqlStyle,
  backgroundcolor=\color{csConfirmColor!5}]
SELECT s.Phone FROM frpm f
JOIN schools s ON f.CDSCode=s.CDSCode
WHERE f."Charter School (Y/N)"=1
  AND f."Charter Funding Type"
      ='Directly funded'
  AND s.OpenDate > '2000-01-01';
\end{lstlisting}
\end{csConfirmBox}

\end{tcolorbox}
\end{minipage}%
\hfill
\begin{minipage}[t]{0.485\textwidth}
\begin{tcolorbox}[
  enhanced, breakable,
  colback=csProposeColor!4, colframe=csProposeColor,
  boxrule=0.6pt, arc=2pt,
  title={\small\textbf{Schema Prefill}%
    \normalfont\small\ (Qwen3-4B,\ 4 turns)},
  coltitle=black,
  attach boxed title to top left={yshift=-2pt, xshift=4pt},
  boxed title style={
    colback=csProposeColor!20,
    colframe=csProposeColor,
    boxrule=0.4pt, arc=1pt,
  },
  left=3pt, right=3pt, top=6pt, bottom=2pt,
]

\begin{csTurnBox}
\textbf{\small T1}\enspace\csExploreTag\
{\footnotesize\color{gray}(synthetic prefill)}
\begin{lstlisting}[style=csResultStyle]
CREATE TABLE frpm (CDSCode TEXT,
  `Charter School (Y/N)` INTEGER,
  `Charter Funding Type` TEXT, ...)
CREATE TABLE schools (CDSCode TEXT,
  Phone TEXT, OpenDate DATE, ...)
\end{lstlisting}
\end{csTurnBox}

\begin{csTurnBox}
\textbf{\small T2}\enspace\csProposeTag
\begin{lstlisting}[style=csSchemaStyle]
tables: [schools, frpm]
schools:[CDSCode, Phone, OpenDate]
frpm:   [CDSCode,Charter School (Y/N)]
join:   schools.CDSCode=frpm.CDSCode
\end{lstlisting}
\end{csTurnBox}

\begin{csTurnBox}
\textbf{\small T3}\enspace\csGenerateTag
\begin{lstlisting}[style=csSqlStyle]
SELECT DISTINCT s.Phone
FROM schools s
INNER JOIN frpm f
  ON s.CDSCode = f.CDSCode
WHERE f."Charter School (Y/N)" = 1
  AND s.OpenDate > '2000-01-01'
\end{lstlisting}
\csResultOmit
\end{csTurnBox}

\begin{csConfirmBox}{T4}
\begin{lstlisting}[style=csSqlStyle,
  backgroundcolor=\color{csConfirmColor!5}]
SELECT DISTINCT s.Phone
FROM schools s
INNER JOIN frpm f
  ON s.CDSCode = f.CDSCode
WHERE f."Charter School (Y/N)" = 1
  AND s.OpenDate > '2000-01-01'
\end{lstlisting}
\end{csConfirmBox}

\end{tcolorbox}
\end{minipage}

\caption{Case study on BIRD-Dev instance \texttt{dev\_4}
  (database \texttt{california\_schools}, Qwen3-4B, greedy decoding).
  \textbf{Left:} Unknown Schema setting with interactive metadata exploration.
  \textbf{Right:} Schema Prefill variant where the complete schema
  is injected as a synthetic \textit{Explore} turn.}
\label{fig:case_study}
\end{figure*}

\end{document}